\documentclass[sigconf]{acmart}

\usepackage{subcaption}

\AtBeginDocument{%
  \providecommand\BibTeX{{%
    \normalfont B\kern-0.5em{\scshape i\kern-0.25em b}\kern-0.8em\TeX}}}

\setcopyright{acmlicensed}
\copyrightyear{2024}
\acmYear{2024}
\acmDOI{XXXXXXX.XXXXXXX}

\acmISBN{978-1-4503-XXXX-X/18/06}

\begin{document}

\title{Self-Attention Mechanism in Multimodal Context for Banking Transaction Flow}

\author{Cyrile Delestre}
\affiliation{%
  \institution{Crédit Mutuel Arkéa}
  \city{Le Relecq-Kerhuon}
  \country{France}
  }

\author{Yoann Sola}
\affiliation{%
  \institution{Crédit Mutuel Arkéa}
  \city{Le Relecq-Kerhuon}
  \country{France}
  }

\renewcommand{\shortauthors}{Delestre and Sola}

\begin{abstract}
  Banking Transaction Flow (BTF) is a sequential data found in a number of banking activities such as marketing, credit risk or banking fraud. It is a multimodal data composed of three modalities: a date, a numerical value and a wording. We propose in this work an application of self-attention mechanism to the processing of BTFs. We trained two general models on a large amount of BTFs in a self-supervised way: one RNN-based model and one Transformer-based model. We proposed a specific tokenization in order to be able to process BTFs. The performance of these two models was evaluated on two banking downstream tasks: a transaction categorization task and a credit risk task. The results show that fine-tuning these two pre-trained models allowed to perform better than the state-of-the-art approaches for both tasks.
\end{abstract}





\received{09 February 2024}
\received[revised]{TBD}
\received[accepted]{TBD}

\maketitle

\section{Introduction}

Machine learning (ML) in the banking system has been a growing practice in recent years and can be found in all its related activities. Banking Transaction Flows (BTF) are often used in customer-related subjects, because it is an important data containing a certain amount of information about the customer himself, which is by nature extremely revealing and difficult to falsify.

The two main banking areas where BTF are used are undeniably marketing and risk.\\
\textbf{Marketing:} In a marketing context, the information we want to extract from banking transactions is the type of spending habits or the household income. This information allows us to advise or categorize consumers according to a commercial or customer knowledge objective. Nowadays, most banks offer Personal Financial Management (PFM), which can help the clients to improve financial management, through a personalized view of finances and advice, without having any knowledge.\\
Marketing segmentation is also a task where banking operations can be very useful. Segmentation is a central discipline in the commercial strategy of a company (not only banking) \cite{smith56}, in which banking flows can be used to extract information on customer knowledge, allowing a better understanding of customer behavior and making targeting more relevant\cite{smeureanu13}.\\
\textbf{Risk:} Since the last global financial crises, risk management has become extremely regulated and monitored in the banking sector \cite{basel06}. The purpose of this risk monitoring is to limit the systemic financial risk \cite{billio12} and thus preserve the integrity of the national and global banking system. Among the various forms of banking risk \cite{leo19}, credit risk is a major one. The bank solvency criterion is nowadays very closely followed by the regulatory and prudential agencies. For the bank, it can be summarized as determining a credit risk and a trade-off between commercial and prudential strategy. In this context, determining a credit default score at the time the credit is contracted is a way to choose the credit risk exposure. The translation of the commercial/prudential trade-off is often expressed by an acceptance threshold defined on the calculated risk score. For this application, banking transactions are widely used and allow to extract insights such as financial health, saving capacity, household spending habits.

In the context of Open Banking \cite{Brodsky17}, BTFs can be exchanged between banks or private/public organizations in order to provide more financial services to their respective customers. These exchanges are rigorously framed by the PSD2 (Payment Services Directive 2) \cite{Regulation15} and by the GDPR (General Data Protection Regulation) \cite{Regulation16}. Using this data standardization, we aim to train a model able to process PSD2-based BTFs. Such a generic pre-trained could be useful for a great number of organizations.

In a lot of use cases, the information encapsulated by BTFs is not always fully exploited. BTFs are often transformed via a feature extraction phase (\textit{e.g.}, incomes estimation, counting the transactions number, etc) and the different modalities of the data are not always kept. A part of the information is lost, as well as the sequential nature of the data. In this work, we aim to process BTFs more efficiently by keeping its multimodal and sequential nature.

We will begin by describing the BTF data as well as the preprocessing we performed. The tokenization phase is one of our main contributions and will be extensively discussed, before describing the two modelling approaches we chose: Recurrent Neural Network (RNN) and Transformer. The self-attention mechanism is a key feature of these models. Another contribution we propose is the design of the pre-training process: we defined several subtasks specific to the multimodal nature of BTFs. We also carried out a hardware performance study, as well as an evaluation of the two pre-trained models on two different downstream tasks: a transaction categorization task and a credit risk task.

\section{Related Work}

We found several application of machine learning to banking activities in the literature, such as marketing segmentation \cite{shaw01}\cite{smeureanu13}, general banking risk management \cite{leo19} and credit risk \cite{galindo00}\cite{bellotti09}\cite{hamori18}. Operational risks were also dealt by several works, \textit{e.g.} the detection of fraudulent transactions \cite{wiese09}\cite{pun12} or money laundering \cite{kute21}.\\
Some works successfully tried to used machine learning in temporal point processing \cite{yan2019recent}\cite{shchur2021neural}, allowing to model event sequences in continuous time space.\\
We also found interesting research about deep learning applied to BTF modelling: \cite{10.1145/3514221.3526129} used contrastive learning inside a self-supervised learning process, \cite{10.1145/3292500.3330693} applied RNN for a credit loan use case, and \cite{10.1145/3447548.3467145} made adversarial attacks on deep learning models of transaction records. These works are not based on the same framework as our work (the PSD2 framework), and often include more features than our BTF definition, \textit{e.g.} the Merchant Category Code (MCC).

The self-attention mechanism first appeared in the Natural Language Processing (NLP) field \cite{bahdanau2015neural}: the words are decomposed in tokens (\textit{e.g.} subwords) and the attention allows to indicate the semantic links between all the tokens from a given sequence (a sentence or a paragraph). Each token is processed with respect to the context around it (\textit{i.e.} all the tokens before and after). The attention mechanism quantifies the relationship between events within a given sequence (the model is then called an encoder) or between two sequences (a cross-encoder). In the literature, it can be found associated with RNN \cite{liu2020multi} or inside the Transformer architecture \cite{vaswani17}.\\
The attention mechanism also appeared in several other fields such as image processing \cite{dosovitskiy2021an} or audio processing \cite{gong21b_interspeech}. It also started to be used in banking use cases: in credit card fraud detection \cite{benchaji2021enhanced}, in credit risk \cite{electronics12071643}, in stock price prediction \cite{doi:10.1080/14697688.2019.1622287}, as well as for general representation of BTF \cite{10.1145/3487553.3524643}.\\
All these works are promising and shows a real interest of the deep learning community in the banking areas. However, we did not find deep learning modelling approaches based on the PSD2 definition of BTF. We will see that the use of self-attention mechanism can fulfill these need, allowing to create a useful generic model in the context of open banking.

\section{Banking Transaction Flow}

BTF represents all the events of banking transactions and a transaction is an event carried out on a bank account of a natural person (as opposed to a legal person). This transaction represents a bank transfer, a withdrawal from an Automated Teller Machine (ATM), a check issue or remittance, a purchase from a Point of Sale (POS), \textit{etc.} The scope defined in the PSD2 framework is the events set that occur on the current account (also called checking account). An event is represented by three modalities:
\begin{enumerate}
   \item The transaction processing date is the date the transaction was taken into account and is officially reflected in the customer's account maintenance. The date is only accurate to the day and, depending on the channel through which the event transited, may have a one or two day delay between the action taken by the customer and the official presence on his account. This date represents the first modality and offers information on the chronology of events on a monthly scale (the year scale will not be taken into account in this paper). These are therefore macro-ordered events but disordered in the daily temporality;
   \item The second modality is the amount associated to the transaction, which represents the transaction value. This is a real number and the sign indicates the transaction direction (debit or credit). Hereafter, this value is in euros, but it can be in any other unit;
   \item Finally, the third and last modality is the wording that accompanies the transaction. This is rich information that indicates the transaction channel (ATM, check, etc.) but also includes information that may be of personal origin (wording instructed by the client for debit transfers, for example) or organizational (wording instructed by a third party for credit transfers or purchases via a POS, for example).
\end{enumerate}

\begin{figure*}[ht]
    \vskip -0.1in
    \centering
    \includegraphics[scale=0.75]{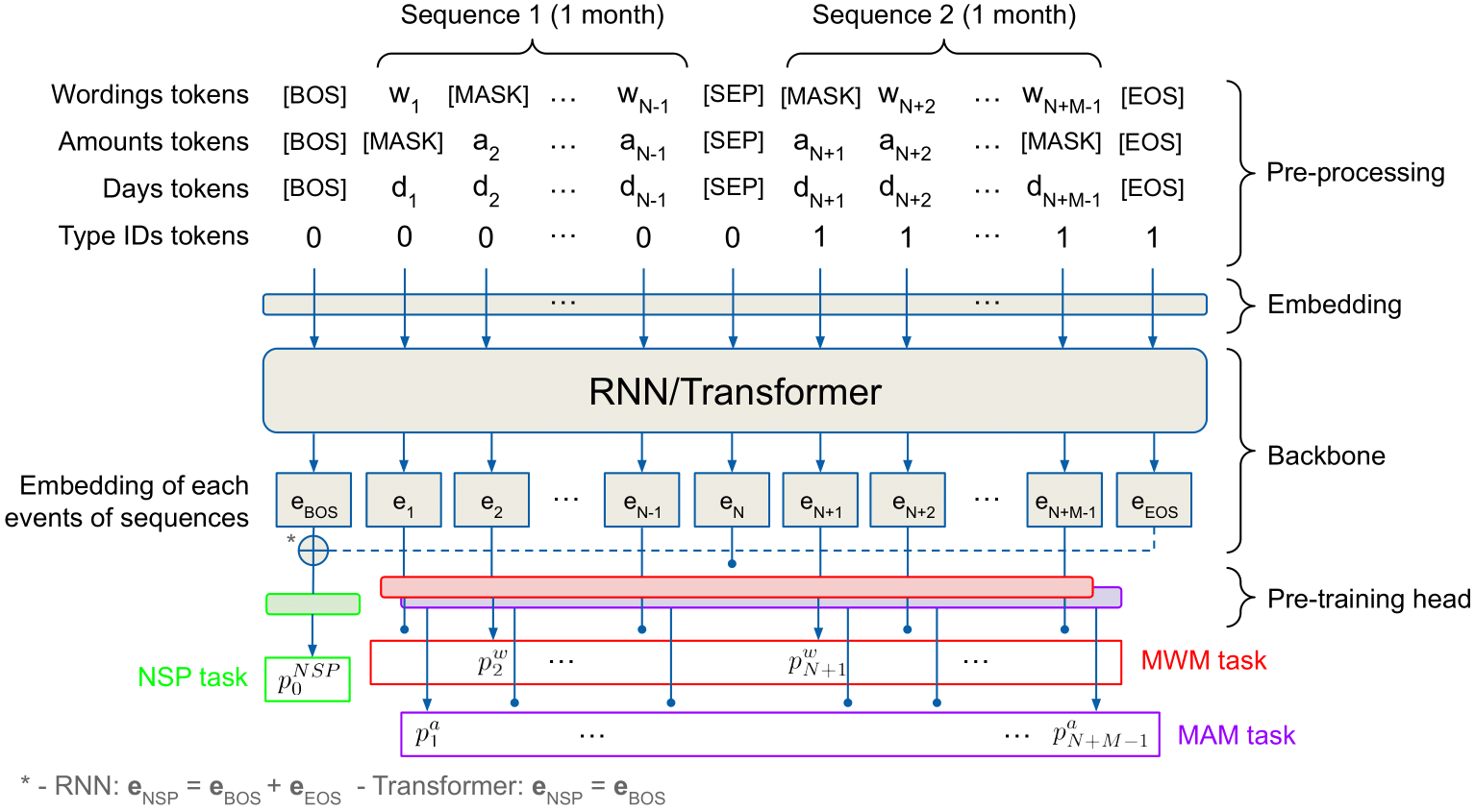}
    \vskip -0.15in
    \caption{Global models diagram and their pre-training heads.}
    \label{fig:structure-training}
    \Description[<short description>]{<long description>}
    \vskip -0.1in
\end{figure*}

In the following section we will present the preprocessing and transformations done to convert a multi-modals events serie into a sequence compatible with and processable by our two models.\\
The figure \ref{fig:structure-training} shows a summary of the whole approach: the tokenization, the models and the sub-tasks of the pre-training process. It should clarify the explanations made throughout the following sections.

\subsection{Preprocessing}\label{normalization}

It is important to carry out a preprocessing phase on this type of data so that it is standardized in order to be robust, efficient and relevant to the modelling we will discuss in the following sections. Moreover, particular attention will be paid to the textual modality of the wordings. Indeed, the latters are free fields and are therefore unnormalized.\\
The wordings have a lot of internal variability that is non-informational or brings a lot of noise, such as check or ATM withdrawal numbers, or even irrelevant information, such as dates, information already carried by the transaction date modality. These parts will be replaced by tags to greatly reduce the non-informational wordings diversity. Also, so that the wordings are not case sensitive, all the characters will be put in lower case and special characters and accents will be removed. Table \ref{tab:preprolib} shows some real examples of what can be found as wordings and their associated normalizations.

\begin{table*}[htbp]
    \centering
    \begin{small}
    \begin{tabular}{ |c|c|c| } 
        \hline
        \textbf{Type} & \textbf{Raw data} & \textbf{Preprocessed data} \\
        \hline
        Digits & \begin{tabular}{c} 
            CHQ 2141367\\
            RET DAB 351267 PLANCOET
        \end{tabular} & \begin{tabular}{c} 
            chq $<$digits$>$\\
            ret dab $<$digits$>$ plancoet
        \end{tabular} \\ 
        \hline
        Date & \begin{tabular}{c}
            VIR POLE EMPLOI BRETAGNE 08/21\\
            CARTE 08/10 LECLERC BREST
        \end{tabular} & \begin{tabular}{c}
            vir pole emploi bretagne $<$date$>$\\
            carte $<$date$>$ leclerc brest
        \end{tabular}\\ 
        \hline
        Other & \begin{tabular}{c}
            @!\_
        \end{tabular} & \begin{tabular}{c}
            $<$empty$>$
        \end{tabular}\\
        \hline
    \end{tabular}
    \end{small}
    \caption{Pattern detection and wording normalization.}
    \label{tab:preprolib}
\end{table*}

\subsection{Tokenization}

A key phase of the modeling is to build the morphosyntax of BTF. That is to say, building a syntax and a dictionary appropriate to the events and to the sequence of all these events. The adopted strategy is first of all a daily ordering of the events by amounts ascending order. In this way, the amounts embedding representation will guarantee the intra-daily position encoding and the amount information.\\
So that an event is consistent with respect to the tokenization, it remains to treat the case of the ``space" character. Indeed, we notice that once the events are juxtaposed to the others, there are two separation types, the extra-wording spaces symbolizing the separation between two events, and the intra-wording spaces symbolizing the words separation inside the wording. To account for this specificity, two different encodings are needed for these two separation types. Thus, if we denote the extra-wording separator by $\Box$ and the intra-wording separator by $\diamond$, and taking the examples from table \ref{tab:preprolib} and assuming that it is already correctly ordered, the wordings sequence gives:\\ 

\begin{figure}[ht]
    \centering
    \vskip -0.1in
    \includegraphics[scale=0.68]{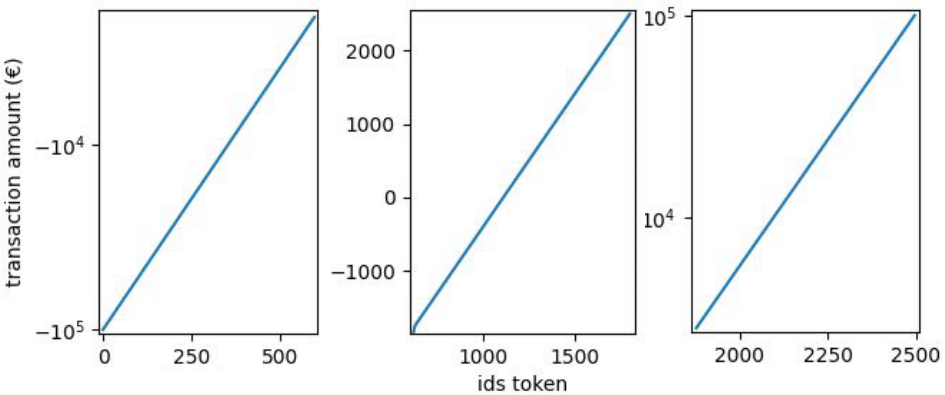}
    \vskip -0.15in
    \caption{Scheme of the three quantizers composing the tokenizer of the amounts. Transaction amount is represented as a function of ids tokens or steps.}
    \label{fig:amount}
    \Description[<short description>]{<long description>}
    \vskip -0.05in
\end{figure}

``\textit{$\Box\diamond$chq$\diamond$$<$digits$>$$\Box\diamond$ret$\diamond$dab$\diamond$$<$digits$>$$\diamond$plancoet \\
$\Box\diamond$vir$\diamond$pole$\diamond$emploi$\diamond$bretagne$\diamond$$<$date$>$ \\
$\Box\diamond$carte$\diamond$$<$date$>$$\diamond$leclerc$\diamond$brest$\Box\diamond$$<$empty$>$}";\\

where in this study we will use as encoding: $\Box = $ U+2581 and $\diamond = \varnothing$.\\
The choice, arbitrary, to choose no encoding for the intra-wording separator character was motivated by the will to more easily attach the company names composed of several words. Thus, the tokenizer will not be more ``influenced" by a character intervening in front of and behind any other character, consequently supporting the creation of more ``independent" atoms (made up of much less composed words).\\
In order to create the dictionary $\mathcal{X}$ of BTF wording, while preserving the encoding specificity of the intra and extra wording separator characters, the SentencePiece Unigram algorithm was chosen \cite{kudo18}, \cite{kudoRicharson18}. Although T. Kudo and J. Richardson \cite{kudoRicharson18} did not note any significant performance difference between a Byte-Pair-Encoding (BPE) \cite{sennrich16} and the Unigram, recent work \cite{zhang20} revealed that a Unigram tokenizer has a better behavior on corpora not dealing with the same information, showing a more generalizing aspect of the created dictionary. This behavior seems to be interesting in the case where the bank flow comes from another organization than the one in which the modeling was not trained. The chosen dictionary size is 7k words and the dictionary has been trained on 1 million sequences composed of 1 month of banking operations.\\
The wording encoding will drive the other tokinizers. For that, we introduce 3 control tokens that will be used later: [BOS] to mark the beginning of a sequence, [EOS] marking the end of a sequence and [SEP] to mark the separation between two sequences. The latter will be used afterwards to contextualize a Natural Language Inference (NLI) problem as a separation marker between a ``premise" sequence and a ``hypothesis" sequence. Thus, we have two cases, mono-sequence and bi-sequence, where here are the possible schemes:
\begin{eqnarray}
   \mathrm{monoSeq} \!\! &=& \!\! [\mathrm{BOS}]\;\mathrm{Seq}\;[\mathrm{EOS}] \label{eq:mono-seq}\\
   \mathrm{biSeq} \!\! &=& \!\! [\mathrm{BOS}]\;\mathrm{Seq}_1\;[\mathrm{SEP}]\;\mathrm{Seq}_2\;[\mathrm{EOS}] \label{eq:bi-seq}
\end{eqnarray}
The amounts tokenizer is composed of a dictionary $\mathcal{A}$ of 2.5k elements. This tokenizer is composed of 3 quantifiers divided into three quantization zones: a linear zone and two exponential zones. In order to create these quantizers and to make them representative of a certain ground reality, 1 million operations amounts were taken randomly to define the different zones boundaries. The linear quantizer extrema were chosen at the first amounts quantile for the low boundary and at the 99th quantile for the high boundary. These limits represent -1750 and +2760 euros and are quantized on 1250 quantization steps. In order to limit the tokenizer saturation, two quantizers exponential composed of 625 steps have been placed at the extremities of the linear quantizer going to the extrema values of saturation chosen at $\pm$100k euros. Figure \ref{fig:amount} illustrates the amounts tokenizer composition. In order to remain coherent with the wordings tokenizer and that the amounts information is reflected on the entire wordings, the amount associated to a transaction will be repeated throughout the decomposition in sub-elements of the wordings tokenization.

For the date modality, a normalization of the day in the month is performed, which allows to bypass the heterogeneity of the month lengths. Then, a linear quantizer allows to quantize the day in the month on 30 quantization steps representing a dictionary size (named $\mathcal{D}$) of 30 elements. Finally, as for the amount tokenizer, the elements will be repeated along the decomposition of the wordings into sub-wordings.

Finally, a last tokenizer is used to mark the identity of the sequence in an NLI context. Indeed, if the sequence is the premise then it will be marked by a token identity of 0 and, if it represents the hypothesis it will be marked by the identity 1. In the mono-sequence case (eq.\ref{eq:mono-seq}) all the elements will have an identity of 0.

To conclude this part, a representation of all these preprocessing and tokenization elements are represented in figure \ref{fig:structure-training} in the part upper labelled ``pre-processing". A complete example of preprocessing and tokenization of a raw data is detailed in the Appendix \ref{append:prepro}. In the next section we will discuss in more detail the two models used in this article.

\section{Modelization}

In this part we will discuss the technical aspect of the two models. Firstly, the embedding part allowing to encode the tokens is identical for both models. The goal of this step is to create a latent representation of tokens. The input of the embedding step is a token and the output is a vector with $d$ dimensions. The dimension of the embedding representation is the same for all the modalities. In the following, we will consider that a sequence is composed of $N$ events. We note respectively $\mathbf{X}$, $\mathbf{A}$, $\mathbf{D}$ and $\mathbf{T}$ all in $\mathbf{R}^{d\times N}$, the latent representation of the tokens sequence composing the wordings, the amounts representation, the temporal modality (representing days) and the identity.\\
At each sequence event, only the wordings tokens are different. In order to carry the information in a uniform way from each modality to each event, the final embedding representation will be the sum of each modality contributions. Thus, for a wording decomposed into sub-wordings, the other modalities will add a common bias to all events composing it. We note this representation as follows:
\begin{eqnarray}
    \mathbf{E}_0  \!\!\!\!&=&\!\!\!\! \mathbf{X} + \mathbf{A} + \mathbf{D} + \mathbf{T}\\
    \mathbf{E}_0  \!\!\!\!&=&\!\!\!\! \left[\mathbf{e}_{0,0}, \cdots, \mathbf{e}_{0,n}, \cdots, \mathbf{e}_{0,N-1}\right] \in \mathbb{R}^{d\times N}
\end{eqnarray}
Finally, the parameters number in the modeling embedding part can be summarized as follows:
\begin{equation}
    p_{emb} = d\times\left(\vert\mathcal{X}\vert+\vert\mathcal{A}\vert+\vert\mathcal{D}\vert+2\right) \label{eq:emb}
\end{equation}

\subsection{Recurrent Neural Network}

The first model is an implementation of the historical architecture for this type of problem, \textit{i.e.} RNN. It is built around a bidirectional recurrent network architecture \cite{schuster97} and the final embedded representation is inspired by the ELMo modeling \cite{peters18}. The bidirectional layers allows to get rid of the events causality, so the $n$-th event will be influenced by the events preceding and following it.\\

The RNN structure used is a Long Short Term Memory (LSTM) \cite{hochreiter97} with a projection allowing to have a recurrent network with an internal latency representation larger than the output one. This strategy has already shown some effectiveness in some applications such as speech recognition \cite{sak2014}. The internal dimension of the recurrent model is denoted $h$ and the external one is our event representation dimension $d$ with $h>d$.\\
Both directions of the LSTM are composed of \textit{L} layers and a layer normalization \cite{ba16} is applied to each layer. The representations computed by the two directions are then given to an attention layer. The goal of this layer is to compute the relations between each token representation, using the self-attention mechanism \cite{bahdanau2015neural}.\\

This approach introduced by ELMo \cite{peters18} has two advantages. Firstly, it limits the vanishing gradient effect through the network layers and secondly, it allows to have a model which will make each embedding representation of the network directly contributing to the output. And since these contributions are made of trainable parameters, the model will adapt to the downstream task and choose the optimal abstraction level of representation for this task. Indeed, the lower the layer level, the less the interactions between the different events are taken into account and \textit{vice versa}. In the field of NLP it has already been shown that, for some tasks, the abstraction level can play an important role in the task performance \cite{Jawahar19}.\\
Finally, the parameters number of the RNN modeling part can be written as:
\begin{equation}
    p_{rnn} = 2L\times (9dh + 8h + 2d)+L+1 \label{eq:rnn}
\end{equation}

\subsection{Transformer}

The Transfomer architecture introduced in \cite{vaswani17} made an original use of the attention mechanism, allowing to remove several limitations. Until this architecture, the events ordering was preserved and was primordial for the RNN-based architectures. The information of the ordering is no more necessary for the Transformer model as this neural network architecture is articulated around a functional memory (based on the attention mechanism) that will essentially react according to the presence or absence of events. The ordering becomes secondary and relations between distant events are easier on this type of structure.\\

We used the classical Transformer architecture, with a number \textit{L} of layers. Each layers is composed of two sub-layers:
\begin{itemize}
    \item A multi-head attention layers allowing to compute the relations between the tokens representations in different sub-space: a different vector space is used for each head, learnt during the training process. This sub-layer is composed of \textit{J} heads.
    \item A feed-forward neural network composed of one hidden layer with a dimension \textit{h}, with $h>d$. Contrary to the original paper, the activation function is a GELU \cite{hendrycks26}, more efficient during the learning phase than a ReLU activation. This activation function is relatively common for this type of modeling, \cite{devlin19}, \cite{yinhan19}. A layer normalization \cite{ba16} is also applied to this neural network.
\end{itemize}
The output of both sub-layers is sequence of vectors of dimension \textit{d}.\\

Finally, as for the embedding and RNN parts, here are the parameters number of the Transformer modeling:
\begin{equation}
    p_{tf} = L\times (4d^2 + 2dh + 9d + h) \label{eq:tf}
\end{equation}

\section{Pre-training}

In this section we will discuss the architecture parameters of the two models, the cost function used for training, and the training parameters.

In table \ref{tab:attribute} are specified the models parameters that remained free until then and that allow to define the final topology. The layers number $L$ of the RNN network is the one used in the ELMo architecture \cite{peters18}, the other parameters are inspired by BERT architecture \cite{devlin19}. Thus, using equations \ref{eq:emb}, \ref{eq:rnn} and \ref{eq:tf}, it is possible to determine the parameters number for each of these two networks. It is interesting to note that the parameters number is very close between the two models, representing an equivalent complexity level. Thus the differences in performance measurements cannot be attributed to one model being more complex than the other.\\
The pre-training strategy consists in 3 subtasks that we will detail:\\
\textbf{Masked Wording Model (MWM):} is similar to BERT's Masked Language Model (MLM) but focused on transaction wordings. It consists in training the model to estimate the dynamically masked wording during the training phase. The probability $p_\mathrm{MWM}$ represents the proportion of hidden wordings in the training sequences. Masking is done at the wording level and not at the subwordings level in the tokenization output.\\
\textbf{Masked Amount Model (MAM):} which consists in estimating the dynamically masked amounts. The probability $p_\mathrm{MAM}$ symbolizes the proportion of the masked amounts and the amounts estimation is done at the wording level. Thus, if an amount a associeted to a wording is masked, the mask is repeated as many times as the wording is broken down by the tokenization.\\
\textbf{Next Sequence Prediction (NSP):} this NLI task is similar to the Next Sentence Prediction (NSP) task in BERT. This task defines the sequences strategy (\textit{e.g.} eq.\ref{eq:bi-seq}) as input to the modeling for pre-training. Thus the first sequence will be one transaction month and the second sequence the continuation of the second month of an individual, or not. This sequence continuity probability will be modeled by $p_\mathrm{NSP}$.\\
\begin{table}[t]
    \begin{center}
        \begin{small}
            \begin{sc}
                \begin{tabular}{r|cc}
                    \toprule
                    Attribute                                               & RNN  & Transformer \\
                    \midrule
                    $d$                                                     & \multicolumn{2}{c}{$768$}\\
                    $h$                                                     & \multicolumn{2}{c}{$3072$}\\
                    $J$                                                     & $\varnothing$ & $12$ \\
                    $L$                                                     & $2$           & $12$\\
                    Layer norm. eps.                                        & \multicolumn{2}{c}{$1\cdot 10^{-5}$}\\        
                    Param. nb. eq.(\ref{eq:emb}, \ref{eq:rnn}, \ref{eq:tf}) & \multicolumn{2}{c}{$92$M}\\
                    \bottomrule
                \end{tabular}
            \end{sc}
        \end{small}
        \caption{Parameters for both structures.}
        \label{tab:attribute}
    \end{center}
    \vskip -0.2in
\end{table}
The loss pre-training function is modeled by the sum of the three subtasks cross-entropies:
\begin{equation}
    \mathcal{L}(p,y) = \mathrm{CE}_{\mathrm{MWM}}(p,y)+ \mathrm{CE}_{\mathrm{MAM}}(p,y)+ \mathrm{CE}_{\mathrm{NSP}}(p,y)
\end{equation}
This is a hard label type of modeling ($y\in\{0,1\}$). It is then possible to define for each of the sub-tasks the ground truth indices:
\begin{equation}
    \mathcal{P}_k = \left\lbrace (i,j):\; i\triangleq\{o_{k,i}\}=\mathcal{O}_k,\; j=\underset{d\in\vert\mathcal{D}_k\vert}{\mathrm{argmax}}\;y_{i,d}\right\rbrace 
\end{equation}
with $k\in \{\mathrm{MWM}, \mathrm{MAM}, \mathrm{NSP}\}$, $\mathcal{O}_k$ the task observation set, respectively $\mathcal{D}_\mathrm{MWM}$ and $\mathcal{D}_\mathrm{MAM}$ the dictionaries $\mathcal{X}$ and $\mathcal{A}$ and $\mathcal{D}_\mathrm{NSP}=\{0, 1\}$. Then the cross-entropy functions can be summarized as:
\begin{equation}
    \mathrm{CE}_{k}(p^k,y^k) = \frac{-1}{\vert \mathcal{O}_k\vert}\sum_{(i,j)\in\mathcal{P}_k}\log(p_{i,j}^k)
\end{equation}
This writing allows to create a simple link between the probability of correct target prediction and the cost function (more details can be found in the Appendix \ref{append:pretrain}). Finally, the parameters used for learning are shown in table \ref{tab:param-learn}. The learning rate strategy improves the training performance on complex networks \cite{smith19}.
\begin{table}[t]
    \begin{center}
        \begin{small}
            \begin{sc}
                \begin{tabular}{r|cc}
                    \toprule
                    Attribute & RNN & Transformer \\
                    \midrule
                    Hardware             & \multicolumn{2}{c}{$8\times$GPU Nvidia A100 $40$GB}\\
                    Set obs./tok. nb.    & \multicolumn{2}{c}{$3$M/$1.29$G $\sim70$GB}\\
                    Max. seq. tok. nb.   & $1500$     & $800$\\
                    Epoch nb.            & \multicolumn{2}{c}{$50$}\\
                    Mini-batch size      & \multicolumn{2}{c}{$32$}\\
                    Gradient accu.       & \multicolumn{2}{c}{$32$}\\
                    $p_\mathrm{dropout}$ & \multicolumn{2}{c}{$0.1$}\\
                    $p_\mathrm{MWM}$     & \multicolumn{2}{c}{$0.05$}\\
                    $p_\mathrm{MAM}$     & \multicolumn{2}{c}{$0.05$}\\
                    $p_\mathrm{NSP}$     & \multicolumn{2}{c}{$0.5$}\\
                    Weight decay         & \multicolumn{2}{c}{$0.01$}\\
                    \multicolumn{1}{r|}{Learning rate}  & \multicolumn{2}{c}{\includegraphics[scale=0.5]{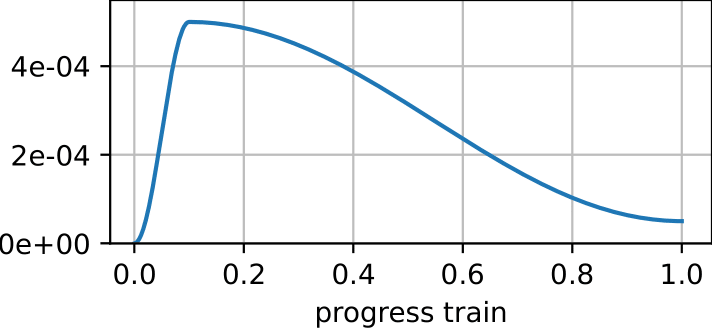}}\\
                    Training time       & 6d 3h37min & 5d 13h18min \\
                \bottomrule
                \end{tabular}
            \end{sc}
        \end{small}
        \caption{Hardware used, pre-training parameters and pre-training times.}
        \label{tab:param-learn}
    \end{center}
    \vskip -0.3in
\end{table}

\section{Performances and Downstream Tasks}

In this experimentation part, we will discuss the models performance in terms of execution time and RAM consumption. We will finish with two downstream tasks on two very different subjects allowing to notice the good behavior of the modeling on very diverse tasks.

\subsection{Hardware Performance}
\begin{table*}[ht]
    \begin{center}
        \begin{small}
            \begin{sc}
                \begin{tabular}{r|ccccc}
                    \toprule
                    model / ($\mathrm{ms}\pm\mathrm{ms}$) & 1 core  & 4 cores & 8 cores & 16 cores & GPU \\
                    \midrule
                    RNN & \footnotesize{$\mathbf{897\pm 126}$} & \footnotesize{$\mathbf{343\pm 37}$} & \footnotesize{$273\pm 30$} & \footnotesize{$274\pm 45$} & \footnotesize{$30\pm 9$}\\
                    Transformer & \footnotesize{$1254\pm 360$} & \footnotesize{$397\pm 97$} & \footnotesize{$\mathbf{226\pm 53}$} & \footnotesize{$\mathbf{149\pm 20}$} & \footnotesize{$\mathbf{17\pm 11}$} \\
                    \bottomrule
                \end{tabular}
            \end{sc}
        \end{small}
        \caption{Average computation time in milliseconds on Intel Xeon CPUs @ 2.2GHz and a Nvidia A100 40GB GPU. The calculation was performed on a sequence of 1 month of bank operations over 500 observations with a mini-batch size of 25 (the calculation time is therefore divided by 25 to return to a sequence level).}
        \label{tab:exe-time}
    \end{center}
    \vskip -0.25in
\end{table*}
The two models have a very different topological nature and therefore different behaviors with respect to parallelization and RAM consumption. Knowing these characteristics allows to better choose the architecture according to the needs and constraints (real time, batch computing, RAM limit, computing power limit, \textit{etc.}).\\
Table \ref{tab:exe-time} summarizes the execution performances of the two models as a function of the computing cores number and on GPU. We can see that with a low parallelization, the RNN architecture is more efficient than Transformer. This may be due to the fact that, in a mini-batch context, padding is calculated for the Transformer architecture whereas in the RNN architecture it is not. As the number transactions variability between sequences can be very large (see Figure \ref{fig:seq-length} in the appendix \ref{append:prepro}) this may explain the advantage of the RNN structure. However, the Transformer structure much better supports parallelization with an inverse linear relationship between the cores number and the execution time. Therefore, on infrastructures with a lot of computing cores, the Transformer structure will be preferred.\\
Though, if the task requires the use of longer sequences (several months for example), the limit in RAM memory may come into account. The RNN structure being recurrent on the events it is thus very little consuming in RAM memory. It nevertheless requires all intermediate layers to calculate the final embedded representation which represents a memory consumption of:
\begin{equation*}
    \mathrm{RNN}\sim dN\times (2L+1) \sim o(N)
\end{equation*}
As for the modeling based on Transformers structures, only the last layer is necessary for the sequence embedding representation. However, internally the attention head operator is much more consuming in RAM, in particular due to the matrix product $\mathbf{Q}^T\mathbf{K} \in \mathbb{R}^{N\times N}$. So we could summarize the consumption of this structure by:
\begin{equation*}
    \mathrm{Transformer}\sim  dN+N^2 \sim o(N^2)
\end{equation*}
We can see that for small sequences the RNN structure will consume more RAM. However the relation between the sequence size and the consumption remains linear whereas for the Transformer structure it is in power 2. So we have two possible strategies: either for tasks requiring long sequences we will prefer the RNN structure or we will choose a sequence truncation appropriate to the amount of RAM memory.

\subsection{Downstream Tasks}

\begin{figure}[ht]
    \centering
    \includegraphics[scale=0.55]{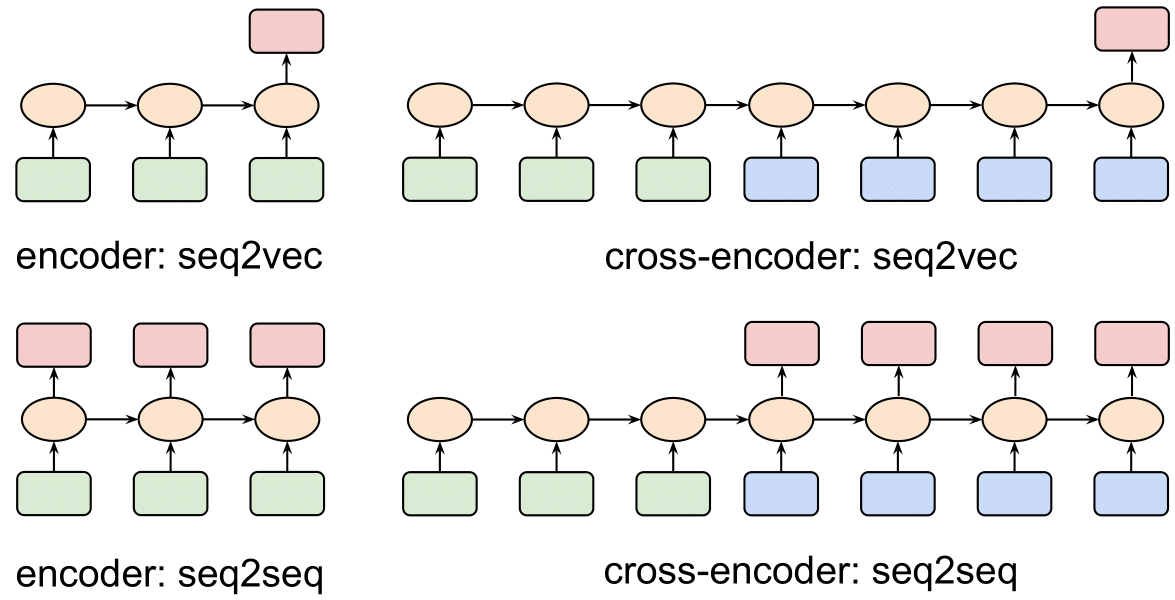}
    \vskip -0.15in
    \caption{Structure type with an encoder. The green (resp. blue) boxes represent the first (resp. second) sequence, the ovals the attention process and the red boxes the output of the models.}
    \label{fig:downstreamtask}
    \Description[<short description>]{<long description>}
    \vskip -0.10in
\end{figure}
Figure \ref{fig:downstreamtask} summarizes the set of processes that can be performed with an encoder. Each of these categories represents one downstream task type that we discussed in the introduction. For example, in the encoder case seq2vec we find the credit risk, seq2seq the operations categorization. In this part we will not discuss any cross-encoder. Indeed, in this context, seq2vec can correspond to identity theft detection that we have already indirectly dealt with because it essentially corresponds to the pre-training NLI subtask (see Figure \ref{fig:train-curve} in the Appendix \ref{append:pretrain}). As for the seq2seq cross-encoder, this structure corresponds in NLP to Question-Answer type tasks. It is quite possible to treat this type of structure, but we have no idea of banking application yet.\\
Faced with the complexity of certain architecture that can be made up of several models and having, in our two cases treated here, enough evaluation observation to satisfy the CLT conditions, it is possible to evaluate the uncertainties thanks to the Normal approximation interval at 95\% (for the accuracy and recall measures). For the ROC-AUC confidence interval in our second use case, the interval expression expressed by \cite{hanley82} will be preferred.\\
Finally, for the two downstream tasks studied in this article, the model input sequence will be composed of 2 months of BTF history. Indeed, in the PSD2 framework, no history depth is imposed to the banks. We have therefore chosen the minimum, allowing us to extract recurrent information between months.\\

\begin{figure}[ht]
    \vskip -0.05in
    \centering
    \includegraphics[scale=0.6]{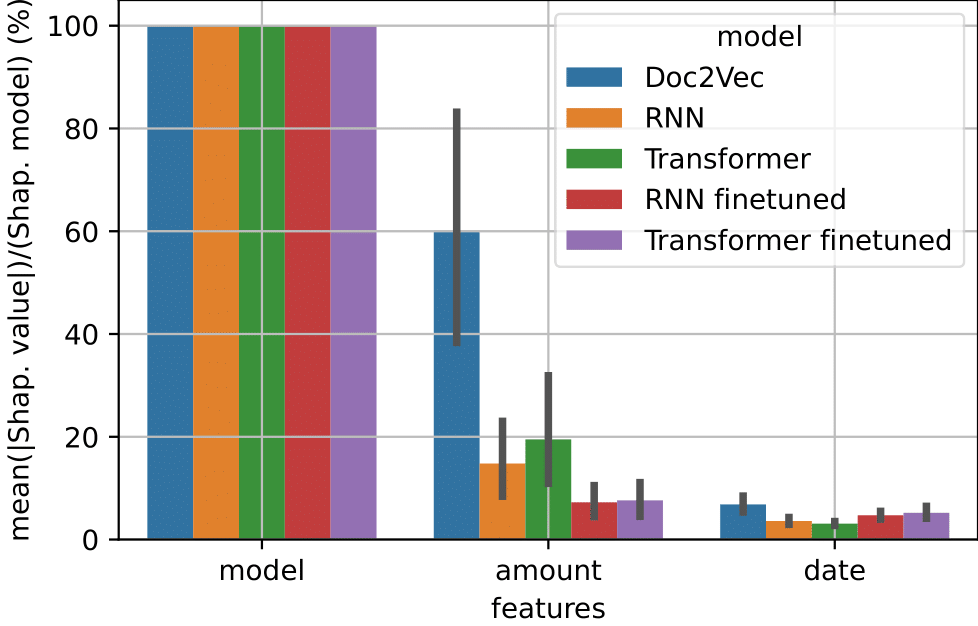}
    \vskip -0.2in
    \caption{Contribution impact of each modalities.}
    \label{fig:shap}
    \Description[<short description>]{<long description>}
    \vskip -0.1in
\end{figure}

\textbf{Transaction categorization:} in this first task we will evaluate the models ability to label bank transactions using the categorizations made by the internal PFM as a reference. As we saw in the introduction, the PFM is in charge of categorizing transactions for account management purposes. However, the categorization system represents an imposing IT architecture (large databases, powerful calculation servers) offering little portability to the tool. It remains interesting to have a model that allows the use of the categorization system according to the need. With this in mind, we will compare our models to a Doc2Vec type approach \cite{le14} pre-trained on 200k bank wordings after normalization (Sec.\ref{normalization}). Doc2Vec allows us to have a embedded representation of each bank wording. In order to integrate the other modalities we add the amount and the month day normalized as input to a Gradient Boosting Decision Tree (GBDT). The complete structure of the modeling approach is detailed in the Appendix \ref{append:transclassif}. Our comparison consists in replacing the Doc2Vec modeling with our direct output modeling and then replacing it again with our modeling adapted to the categorization task after a finetuning phase. The performances are summarized in table \ref{tab:perf-pfm}. We can see that, from scratch, the Transformer model has a much better generalization power than the RNN model by clearly distinguishing itself from the reference Doc2Vec model by an accuracy gain of 10 points. But the important and somewhat unexpected result is that, after a finetuning phase, the two models offer almost identical performances with a gain of 28 points on the accuracy compared to the reference model.\\
It is possible to measure the impact of the proposed multi-modal modeling by visualizing the contribution of each modality to the final GBDT model. For this we use the Shapley value \cite{strumbelj10}, and in the particular case where the model is based on binary decision trees we use \cite{lundberg18} to measure this impact. In figure \ref{fig:shap} we can see that the Doc2Vec model (a model taking into account only one modality) has a strong amounts contribution. Indeed, in addition to the wording content, this modality seems to be very important to determine the transaction type. We also notice that, for our models, the vast majority of the information is contained in the model and that, after finetuning, this aspect is still reinforced. This observation highlights the multi-modal aspect of our models and shows their good exploitation by the models.

\begin{table}[t]
    \begin{center}
        \begin{scriptsize}
            \begin{sc}
                \begin{tabular}{r|ccc}
                    \toprule
                    model / ($\%$)   & Accuracy               & Recall                 & f1-score \\
                    \midrule
                    Doc2Vec          & $62.5\pm 0.9$          & $62.3\pm 0.9$          & $62.3$\\
                    RNN              & $62.0\pm 1.0$          & $62.1\pm 1.0$          & $61.9$\\
                    Transformer      & $76.0\pm 0.8$          & $76.1\pm 0.8$          & $75.8$\\
                    RNN f.t.         & $\mathbf{89.5\pm 0.6}$ & $\mathbf{89.5\pm 0.6}$ & $\mathbf{89.3}$\\
                    Transformer f.t. & $\mathbf{90.4\pm 0.5}$ & $\mathbf{90.4\pm 0.5}$ & $\mathbf{90.2}$\\
                    \bottomrule
                \end{tabular}
            \end{sc}
        \end{scriptsize}
        \caption{Performances on operation categorization task.}
        \label{tab:perf-pfm}
    \end{center}
    \vskip -0.4in
\end{table}

\textbf{Credit risk:} consumer credit risk is a two-class classification task that involves determining a score representing the default risk. It is important to note that a scoring tool to determine credit risk is within the scope of GDPR framework and an explanation must be provided by the banking company to customers who request it on the reasons explaining the score. So, at the company level, it is a trade off between computational cost, XIA and performance. Therefore, the performance gain of a new technology must be significant to justify a paradigm shift. It is up to the company to define what an incidental contract is. In this article, an incident will be defined as any contract that has been at least 15 days late in payment or at least 1 month late during the first year of the contract's life. In order to fit the article subject, the perimeter of the used data will be only the BTF. Other data sources such as socio-demographics, balance amounts, \textit{etc.} that generally increase the predictive quality will not be used.\\
The reference model is a typical model encountered in this problem type, it will first extract the maximum amount of information encapsulated in the BTF (estimation of credit and debit recurrences, savings estimation, fragility detection, \textit{etc.}) defining the model input characteristics. Here the GBDT is well suited for this problem type \cite{hamori18}. We also added a naive deep learning model attempting to jointly exploit the three modalities of the data in order to illustrate the task difficulty. Finally our models will be tested in the case where only the head layer is trained (not changing the network parameters) and in a case where the whole network is finetuned. The performances are summarized in table \ref{tab:perf-risk}. First, we notice that the reference model (GBDT) have quite good performances and similar to the state-of-the-art \cite{bellotti09}, \cite{hamori18}. The naive model (DL) shows the difficulty to fully exploit such a data directly. The set of models presented in this paper shows real gains and we can draw the same conclusions as for the previous task. Further details can be found in the Apeendix \ref{append:creditrisk}.

\begin{table}[t]
    \begin{center}
        \begin{scriptsize}
            \begin{sc}
                \begin{tabular}{r|ccc}
                    \toprule
                    model / ($\%$)   & ROC-AUC       & Accuracy      & f1-score \\
                    \midrule
                    GBDT             & $73.8\pm 2.2$          & $67.9\pm 2.1$          & $67.8$\\
                    DL               & $68.4\pm 2.6$          & $63.4\pm 2.4$          & $63.3$\\
                    RNN              & $80.2\pm 2.1$          & $73.1\pm 2.2$          & $73.1$\\
                    Transformer      & $\mathbf{81.8\pm 2.1}$ & $\mathbf{73.7\pm 2.2}$ & $\mathbf{73.4}$\\
                    RNN f.t.         & $\mathbf{83.4\pm 2.0}$ & $\mathbf{76.5\pm 2.1}$ & $\mathbf{76.2}$\\
                    Transformer f.t. & $\mathbf{84.4\pm 1.9}$ & $\mathbf{77.1\pm 2.1}$ & $\mathbf{77.0}$\\
                    \bottomrule
                \end{tabular}
            \end{sc}
        \end{scriptsize}
        \caption{Performances on credit risk task.}
        \label{tab:perf-risk}
    \end{center}
    \vskip -0.5in
\end{table}

\section{Conclusions}

In this work, we were able to evaluate two modeling approaches of the banking transaction flows, based on the attention mechanisms. It allowed to jointly used the 3 modalities of BTFs and to fully exploit the information contained within.\\
We have also demonstrate through 2 downstream tasks the generalization ability of these pre-trained models, showing that they can be deployed on relatively diverse tasks. In the two tested cases, the observed difference in performance is significant enough for us to justify a change of paradigm for applications based on BTF data.\\
We also found that, without finetuning, the Transformer-based modeling is more generalizing than the RNN-based modeling. But after finetuning, both architectures offered roughly equivalent performances.\\

The fact that these generic modeling approaches were trained on PSD2 data allows its use in a great number of organizations and this work fits perfectly in the context of open banking.\\

Given the large size of the two models (their parameters number), we now would like to apply Knowledge Distillation (KD) techniques to reduce the computational and memory costs \cite{hinton15}\cite{sanh18}. The distilled models allow to significantly reduce the use of the IT infrastructures while offering good performances compared to the original models.\\
We will also try to quantize the models \cite{dettmers2022llm}, in both 4-bit and 8-bit resolutions. This approach has shown very promising results in recent works \cite{dettmers2024qlora}\cite{lin2023awq}, allowing to largely reduce the size of the models, while preserving their performance.


\bibliographystyle{ACM-Reference-Format}
\bibliography{bib_paper}


\begin{thebibliography}{55}


\ifx \showCODEN    \undefined \def \showCODEN     #1{\unskip}     \fi
\ifx \showDOI      \undefined \def \showDOI       #1{#1}\fi
\ifx \showISBNx    \undefined \def \showISBNx     #1{\unskip}     \fi
\ifx \showISBNxiii \undefined \def \showISBNxiii  #1{\unskip}     \fi
\ifx \showISSN     \undefined \def \showISSN      #1{\unskip}     \fi
\ifx \showLCCN     \undefined \def \showLCCN      #1{\unskip}     \fi
\ifx \shownote     \undefined \def \shownote      #1{#1}          \fi
\ifx \showarticletitle \undefined \def \showarticletitle #1{#1}   \fi
\ifx \showURL      \undefined \def \showURL       {\relax}        \fi
\providecommand\bibfield[2]{#2}
\providecommand\bibinfo[2]{#2}
\providecommand\natexlab[1]{#1}
\providecommand\showeprint[2][]{arXiv:#2}

\bibitem[Ba et~al\mbox{.}(2016)]%
        {ba16}
\bibfield{author}{\bibinfo{person}{J.~L. Ba}, \bibinfo{person}{J.~R. Kiros}, {and} \bibinfo{person}{G.~E. Hinton}.} \bibinfo{year}{2016}\natexlab{}.
\newblock \bibinfo{title}{Layer Normalization}.
\newblock
\newblock
\urldef\tempurl%
\url{https://arxiv.org/abs/1607.06450}
\showURL{%
\tempurl}


\bibitem[Babaev et~al\mbox{.}(2022)]%
        {10.1145/3514221.3526129}
\bibfield{author}{\bibinfo{person}{Dmitrii Babaev}, \bibinfo{person}{Nikita Ovsov}, \bibinfo{person}{Ivan Kireev}, \bibinfo{person}{Maria Ivanova}, \bibinfo{person}{Gleb Gusev}, \bibinfo{person}{Ivan Nazarov}, {and} \bibinfo{person}{Alexander Tuzhilin}.} \bibinfo{year}{2022}\natexlab{}.
\newblock \showarticletitle{CoLES: Contrastive Learning for Event Sequences with Self-Supervision}. In \bibinfo{booktitle}{\emph{Proceedings of the 2022 International Conference on Management of Data}} (Philadelphia, PA, USA) \emph{(\bibinfo{series}{SIGMOD '22})}. \bibinfo{publisher}{Association for Computing Machinery}, \bibinfo{address}{New York, NY, USA}, \bibinfo{pages}{1190–1199}.
\newblock
\showISBNx{9781450392495}
\urldef\tempurl%
\url{https://doi.org/10.1145/3514221.3526129}
\showDOI{\tempurl}


\bibitem[Babaev et~al\mbox{.}(2019)]%
        {10.1145/3292500.3330693}
\bibfield{author}{\bibinfo{person}{Dmitrii Babaev}, \bibinfo{person}{Maxim Savchenko}, \bibinfo{person}{Alexander Tuzhilin}, {and} \bibinfo{person}{Dmitrii Umerenkov}.} \bibinfo{year}{2019}\natexlab{}.
\newblock \showarticletitle{E.T.-RNN: Applying Deep Learning to Credit Loan Applications}. In \bibinfo{booktitle}{\emph{Proceedings of the 25th ACM SIGKDD International Conference on Knowledge Discovery \& Data Mining}} (Anchorage, AK, USA) \emph{(\bibinfo{series}{KDD '19})}. \bibinfo{publisher}{Association for Computing Machinery}, \bibinfo{address}{New York, NY, USA}, \bibinfo{pages}{2183–2190}.
\newblock
\showISBNx{9781450362016}
\urldef\tempurl%
\url{https://doi.org/10.1145/3292500.3330693}
\showDOI{\tempurl}


\bibitem[Bahdanau et~al\mbox{.}(2015)]%
        {bahdanau2015neural}
\bibfield{author}{\bibinfo{person}{Dzmitry Bahdanau}, \bibinfo{person}{Kyung~Hyun Cho}, {and} \bibinfo{person}{Yoshua Bengio}.} \bibinfo{year}{2015}\natexlab{}.
\newblock \showarticletitle{Neural machine translation by jointly learning to align and translate}. In \bibinfo{booktitle}{\emph{3rd International Conference on Learning Representations, ICLR 2015}}. \bibinfo{publisher}{dblp}, \bibinfo{address}{Schloss Dagstuhl, Leibniz-Zentrum für Informatik, Oktavie-Allee,66687 Wadern, Germany}, \bibinfo{pages}{01--15}.
\newblock


\bibitem[BCBS(2006)]%
        {basel06}
\bibfield{author}{\bibinfo{person}{BCBS}.} \bibinfo{year}{2006}\natexlab{}.
\newblock \bibinfo{booktitle}{\emph{Basel II: International convergence of capital measurement and capital standards: A revised framework--comprehensive version}}.
\newblock \bibinfo{type}{{T}echnical {R}eport}. \bibinfo{institution}{Basel Committee on Banking Supervision}.
\newblock


\bibitem[Bellotti and Crook(2009)]%
        {bellotti09}
\bibfield{author}{\bibinfo{person}{T. Bellotti} {and} \bibinfo{person}{J. Crook}.} \bibinfo{year}{2009}\natexlab{}.
\newblock \showarticletitle{Support Vector Machines for Credit Scoring and Discovery of Significant Features}.
\newblock \bibinfo{journal}{\emph{Expert Systems with Applications}} \bibinfo{volume}{36}, \bibinfo{number}{2} (\bibinfo{year}{2009}), \bibinfo{pages}{3302--3308}.
\newblock


\bibitem[Benchaji et~al\mbox{.}(2021)]%
        {benchaji2021enhanced}
\bibfield{author}{\bibinfo{person}{Ibtissam Benchaji}, \bibinfo{person}{Samira Douzi}, \bibinfo{person}{Bouabid El~Ouahidi}, {and} \bibinfo{person}{Jaafar Jaafari}.} \bibinfo{year}{2021}\natexlab{}.
\newblock \showarticletitle{Enhanced credit card fraud detection based on attention mechanism and LSTM deep model}.
\newblock \bibinfo{journal}{\emph{Journal of Big Data}}  \bibinfo{volume}{8} (\bibinfo{year}{2021}), \bibinfo{pages}{1--21}.
\newblock


\bibitem[Billio et~al\mbox{.}(2012)]%
        {billio12}
\bibfield{author}{\bibinfo{person}{M. Billio}, \bibinfo{person}{M. Getmansky}, \bibinfo{person}{A.~W. Lo}, {and} \bibinfo{person}{L. Pelizzon}.} \bibinfo{year}{2012}\natexlab{}.
\newblock \showarticletitle{Econometric Measures of Connectedness and Systemic Risk in the Finance and Insurance Sectors}.
\newblock \bibinfo{journal}{\emph{Journal of Financial Economics}} \bibinfo{volume}{104}, \bibinfo{number}{3} (\bibinfo{year}{2012}), \bibinfo{pages}{535--559}.
\newblock


\bibitem[Brodsky and Oakes(2017)]%
        {Brodsky17}
\bibfield{author}{\bibinfo{person}{Laura Brodsky} {and} \bibinfo{person}{Liz Oakes}.} \bibinfo{year}{2017}\natexlab{}.
\newblock \showarticletitle{Data sharing and open banking}.
\newblock \bibinfo{journal}{\emph{McKinsey \& Company}}  \bibinfo{volume}{1105} (\bibinfo{year}{2017}), \bibinfo{pages}{01--08}.
\newblock


\bibitem[Chen and Ge(2019)]%
        {doi:10.1080/14697688.2019.1622287}
\bibfield{author}{\bibinfo{person}{Shun Chen} {and} \bibinfo{person}{Lei Ge}.} \bibinfo{year}{2019}\natexlab{}.
\newblock \showarticletitle{Exploring the attention mechanism in LSTM-based Hong Kong stock price movement prediction}.
\newblock \bibinfo{journal}{\emph{Quantitative Finance}} \bibinfo{volume}{19}, \bibinfo{number}{9} (\bibinfo{year}{2019}), \bibinfo{pages}{1507--1515}.
\newblock
\urldef\tempurl%
\url{https://doi.org/10.1080/14697688.2019.1622287}
\showDOI{\tempurl}
\showeprint{https://doi.org/10.1080/14697688.2019.1622287}


\bibitem[Dettmers et~al\mbox{.}(2022)]%
        {dettmers2022llm}
\bibfield{author}{\bibinfo{person}{Tim Dettmers}, \bibinfo{person}{Mike Lewis}, \bibinfo{person}{Younes Belkada}, {and} \bibinfo{person}{Luke Zettlemoyer}.} \bibinfo{year}{2022}\natexlab{}.
\newblock \bibinfo{title}{LLM. int8 (): 8-bit Matrix Multiplication for Transformers at Scale. CoRR abs/2208.07339 (2022)}.
\newblock
\newblock


\bibitem[Dettmers et~al\mbox{.}(2024)]%
        {dettmers2024qlora}
\bibfield{author}{\bibinfo{person}{Tim Dettmers}, \bibinfo{person}{Artidoro Pagnoni}, \bibinfo{person}{Ari Holtzman}, {and} \bibinfo{person}{Luke Zettlemoyer}.} \bibinfo{year}{2024}\natexlab{}.
\newblock \showarticletitle{Qlora: Efficient finetuning of quantized llms}.
\newblock \bibinfo{journal}{\emph{Advances in Neural Information Processing Systems}}  \bibinfo{volume}{36} (\bibinfo{year}{2024}), \bibinfo{pages}{01--28}.
\newblock


\bibitem[Devlin et~al\mbox{.}(2019)]%
        {devlin19}
\bibfield{author}{\bibinfo{person}{Jacob Devlin}, \bibinfo{person}{Ming-Wei Chang}, \bibinfo{person}{Kenton Lee}, {and} \bibinfo{person}{Kristina Toutanova}.} \bibinfo{year}{2019}\natexlab{}.
\newblock \showarticletitle{BERT: Pre-training of Deep Bidirectional Transformers for Language Understanding}. In \bibinfo{booktitle}{\emph{Proceedings of NAACL-HLT}}. \bibinfo{publisher}{Association for Computational Linguistics}, \bibinfo{address}{Minneapolis, Minnesota}, \bibinfo{pages}{4171--4186}.
\newblock
\urldef\tempurl%
\url{https://api.semanticscholar.org/CorpusID:52967399}
\showURL{%
\tempurl}


\bibitem[Dosovitskiy et~al\mbox{.}(2020)]%
        {dosovitskiy2021an}
\bibfield{author}{\bibinfo{person}{Alexey Dosovitskiy}, \bibinfo{person}{Lucas Beyer}, \bibinfo{person}{Alexander Kolesnikov}, \bibinfo{person}{Dirk Weissenborn}, \bibinfo{person}{Xiaohua Zhai}, \bibinfo{person}{Thomas Unterthiner}, \bibinfo{person}{Mostafa Dehghani}, \bibinfo{person}{Matthias Minderer}, \bibinfo{person}{Georg Heigold}, \bibinfo{person}{Sylvain Gelly}, \bibinfo{person}{Jakob Uszkoreit}, {and} \bibinfo{person}{Neil Houlsby}.} \bibinfo{year}{2020}\natexlab{}.
\newblock \showarticletitle{An Image is Worth 16x16 Words: Transformers for Image Recognition at Scale}.
\newblock \bibinfo{journal}{\emph{ArXiv}}  \bibinfo{volume}{abs/2010.11929} (\bibinfo{year}{2020}), \bibinfo{pages}{01--22}.
\newblock
\urldef\tempurl%
\url{https://api.semanticscholar.org/CorpusID:225039882}
\showURL{%
\tempurl}


\bibitem[{European Commission} et~al\mbox{.}(2015)]%
        {Regulation15}
\bibfield{author}{\bibinfo{person}{{European Commission}}, \bibinfo{person}{{Council of the European Union}}, {and} \bibinfo{person}{{European Parliament}}.} \bibinfo{year}{2015}\natexlab{}.
\newblock \bibinfo{booktitle}{\emph{Regulation (EU) 2015/2365 of the European Parliament and of the Council on Transparency of Securities Financing Transactions and of Reuse and Amending Regulation}}.
\newblock \bibinfo{type}{{T}echnical {R}eport} L 337, vol 58. \bibinfo{institution}{Official Journal of the European Union}.
\newblock


\bibitem[{European Commission} et~al\mbox{.}(2016)]%
        {Regulation16}
\bibfield{author}{\bibinfo{person}{{European Commission}}, \bibinfo{person}{{Council of the European Union}}, {and} \bibinfo{person}{{European Parliament}}.} \bibinfo{year}{2016}\natexlab{}.
\newblock \bibinfo{booktitle}{\emph{Regulation (EU) 2016/679 of the European Parliament and of the Council on the Protection of Natural Persons with Regard to the Processing of Personal Data and on the Free Movement of Such Data, and Repealing Directive 95/46/EC (General Data Protection Regulation)}}.
\newblock \bibinfo{type}{{T}echnical {R}eport} L 119, vol 59. \bibinfo{institution}{Official Journal of the European Union}.
\newblock


\bibitem[Fursov et~al\mbox{.}(2021)]%
        {10.1145/3447548.3467145}
\bibfield{author}{\bibinfo{person}{Ivan Fursov}, \bibinfo{person}{Matvey Morozov}, \bibinfo{person}{Nina Kaploukhaya}, \bibinfo{person}{Elizaveta Kovtun}, \bibinfo{person}{Rodrigo Rivera-Castro}, \bibinfo{person}{Gleb Gusev}, \bibinfo{person}{Dmitry Babaev}, \bibinfo{person}{Ivan Kireev}, \bibinfo{person}{Alexey Zaytsev}, {and} \bibinfo{person}{Evgeny Burnaev}.} \bibinfo{year}{2021}\natexlab{}.
\newblock \showarticletitle{Adversarial Attacks on Deep Models for Financial Transaction Records}. In \bibinfo{booktitle}{\emph{Proceedings of the 27th ACM SIGKDD Conference on Knowledge Discovery \& Data Mining}} (Virtual Event, Singapore) \emph{(\bibinfo{series}{KDD '21})}. \bibinfo{publisher}{Association for Computing Machinery}, \bibinfo{address}{New York, NY, USA}, \bibinfo{pages}{2868–2878}.
\newblock
\showISBNx{9781450383325}
\urldef\tempurl%
\url{https://doi.org/10.1145/3447548.3467145}
\showDOI{\tempurl}


\bibitem[Galindo and Tamayo(2000)]%
        {galindo00}
\bibfield{author}{\bibinfo{person}{J. Galindo} {and} \bibinfo{person}{P. Tamayo}.} \bibinfo{year}{2000}\natexlab{}.
\newblock \showarticletitle{Credit Risk Assessment Using Statistical and Machine Learning: Basic Methodology and Risk Modeling Applications}.
\newblock \bibinfo{journal}{\emph{Computational Economics}} \bibinfo{volume}{15}, \bibinfo{number}{1} (\bibinfo{year}{2000}), \bibinfo{pages}{107--143}.
\newblock


\bibitem[Gong et~al\mbox{.}(2021)]%
        {gong21b_interspeech}
\bibfield{author}{\bibinfo{person}{Yuan Gong}, \bibinfo{person}{Yu-An Chung}, {and} \bibinfo{person}{James~R. Glass}.} \bibinfo{year}{2021}\natexlab{}.
\newblock \showarticletitle{AST: Audio Spectrogram Transformer}.
\newblock \bibinfo{journal}{\emph{ArXiv}}  \bibinfo{volume}{abs/2104.01778} (\bibinfo{year}{2021}), \bibinfo{pages}{01--05}.
\newblock
\urldef\tempurl%
\url{https://api.semanticscholar.org/CorpusID:233024831}
\showURL{%
\tempurl}


\bibitem[Hamori et~al\mbox{.}(2018)]%
        {hamori18}
\bibfield{author}{\bibinfo{person}{S. Hamori}, \bibinfo{person}{M. Kawai}, \bibinfo{person}{T. Kume}, \bibinfo{person}{Y. Murakami}, {and} \bibinfo{person}{C. Watanabe}.} \bibinfo{year}{2018}\natexlab{}.
\newblock \showarticletitle{Ensemble Learning or Deep Learning? Application to Default Risk Analysis}.
\newblock \bibinfo{journal}{\emph{Journal of Risk and Financial Management}} \bibinfo{volume}{11}, \bibinfo{number}{1} (\bibinfo{year}{2018}), \bibinfo{pages}{12}.
\newblock


\bibitem[Hanley and McNeil(1982)]%
        {hanley82}
\bibfield{author}{\bibinfo{person}{J.~A. Hanley} {and} \bibinfo{person}{B.~J. McNeil}.} \bibinfo{year}{1982}\natexlab{}.
\newblock \showarticletitle{The Meaning and Use of the Area Under a Receiver Operating Characteristic ({ROC}) Curve}.
\newblock \bibinfo{journal}{\emph{Radiology}} \bibinfo{volume}{143}, \bibinfo{number}{1} (\bibinfo{year}{1982}), \bibinfo{pages}{29--36}.
\newblock


\bibitem[Hendrycks and Gimpel(2016)]%
        {hendrycks26}
\bibfield{author}{\bibinfo{person}{D. Hendrycks} {and} \bibinfo{person}{K. Gimpel}.} \bibinfo{year}{2016}\natexlab{}.
\newblock \bibinfo{title}{Gaussian Error Linear Units ({GELUs})}.
\newblock
\newblock
\urldef\tempurl%
\url{https://arxiv.org/abs/1606.08415}
\showURL{%
\tempurl}


\bibitem[Hinton et~al\mbox{.}(2015)]%
        {hinton15}
\bibfield{author}{\bibinfo{person}{G. Hinton}, \bibinfo{person}{O. Vinyals}, {and} \bibinfo{person}{J. Dean}.} \bibinfo{year}{2015}\natexlab{}.
\newblock \bibinfo{title}{Distilling the Knowledge in a Neural Network}.
\newblock
\newblock
\urldef\tempurl%
\url{https://arxiv.org/abs/1503.02531}
\showURL{%
\tempurl}


\bibitem[Hochreiter and Schmidhuber(1997)]%
        {hochreiter97}
\bibfield{author}{\bibinfo{person}{S. Hochreiter} {and} \bibinfo{person}{J. Schmidhuber}.} \bibinfo{year}{1997}\natexlab{}.
\newblock \showarticletitle{Long Short-term Memory}.
\newblock \bibinfo{journal}{\emph{Neural Computation}} \bibinfo{volume}{9}, \bibinfo{number}{8} (\bibinfo{year}{1997}), \bibinfo{pages}{1735--1780}.
\newblock


\bibitem[Jawahar et~al\mbox{.}(2019)]%
        {Jawahar19}
\bibfield{author}{\bibinfo{person}{Ganesh Jawahar}, \bibinfo{person}{Beno{\^i}t Sagot}, {and} \bibinfo{person}{Djam{\'e} Seddah}.} \bibinfo{year}{2019}\natexlab{}.
\newblock \showarticletitle{What Does BERT Learn about the Structure of Language?}. In \bibinfo{booktitle}{\emph{Annual Meeting of the Association for Computational Linguistics}}. \bibinfo{publisher}{Association for Computational Linguistics}, \bibinfo{address}{Florence, Italy}, \bibinfo{pages}{3651–--3657}.
\newblock
\urldef\tempurl%
\url{https://api.semanticscholar.org/CorpusID:195477534}
\showURL{%
\tempurl}


\bibitem[Kudo(2018)]%
        {kudo18}
\bibfield{author}{\bibinfo{person}{Taku Kudo}.} \bibinfo{year}{2018}\natexlab{}.
\newblock \showarticletitle{Subword Regularization: Improving Neural Network Translation Models with Multiple Subword Candidates}.
\newblock \bibinfo{journal}{\emph{ArXiv}}  \bibinfo{volume}{abs/1804.10959} (\bibinfo{year}{2018}), \bibinfo{pages}{01--10}.
\newblock
\urldef\tempurl%
\url{https://api.semanticscholar.org/CorpusID:13753208}
\showURL{%
\tempurl}


\bibitem[Kudo and Richardson(2018)]%
        {kudoRicharson18}
\bibfield{author}{\bibinfo{person}{T. Kudo} {and} \bibinfo{person}{J. Richardson}.} \bibinfo{year}{2018}\natexlab{}.
\newblock \bibinfo{title}{Sentencepiece: A Simple and Language Independent Subword Tokenizer and Detokenizer for Neural Text Processing}.
\newblock
\newblock
\urldef\tempurl%
\url{https://arxiv.org/abs/1808.06226}
\showURL{%
\tempurl}


\bibitem[Kute et~al\mbox{.}(2021)]%
        {kute21}
\bibfield{author}{\bibinfo{person}{D.~V. Kute}, \bibinfo{person}{B. Pradhan}, \bibinfo{person}{N. Shukla}, {and} \bibinfo{person}{A. Alamri}.} \bibinfo{year}{2021}\natexlab{}.
\newblock \showarticletitle{Deep learning and explainable artificial intelligence techniques applied for detecting money laundering--a critical review}.
\newblock \bibinfo{journal}{\emph{IEEE Access}}  \bibinfo{volume}{9} (\bibinfo{year}{2021}), \bibinfo{pages}{82300--82317}.
\newblock


\bibitem[Le and Mikolov(2014)]%
        {le14}
\bibfield{author}{\bibinfo{person}{Quoc Le} {and} \bibinfo{person}{Tomas Mikolov}.} \bibinfo{year}{2014}\natexlab{}.
\newblock \showarticletitle{Distributed Representations of Sentences and Documents}. In \bibinfo{booktitle}{\emph{Proceedings of the 31st International Conference on Machine Learning}} \emph{(\bibinfo{series}{Proceedings of Machine Learning Research})}, \bibfield{editor}{\bibinfo{person}{Eric~P. Xing} {and} \bibinfo{person}{Tony Jebara}} (Eds.). \bibinfo{publisher}{PMLR}, \bibinfo{address}{Bejing, China}, \bibinfo{pages}{1188--1196}.
\newblock


\bibitem[Leo et~al\mbox{.}(2019)]%
        {leo19}
\bibfield{author}{\bibinfo{person}{M. Leo}, \bibinfo{person}{S. Sharma}, {and} \bibinfo{person}{K. Maddulety}.} \bibinfo{year}{2019}\natexlab{}.
\newblock \showarticletitle{Machine Learning in Banking Risk Management: A Literature Review}.
\newblock \bibinfo{journal}{\emph{Risks}} \bibinfo{volume}{7}, \bibinfo{number}{1} (\bibinfo{year}{2019}), \bibinfo{pages}{29}.
\newblock


\bibitem[Li et~al\mbox{.}(2023)]%
        {electronics12071643}
\bibfield{author}{\bibinfo{person}{Jingyuan Li}, \bibinfo{person}{Caosen Xu}, \bibinfo{person}{Bing Feng}, {and} \bibinfo{person}{Hanyu Zhao}.} \bibinfo{year}{2023}\natexlab{}.
\newblock \showarticletitle{Credit Risk Prediction Model for Listed Companies Based on CNN-LSTM and Attention Mechanism}.
\newblock \bibinfo{journal}{\emph{Electronics}} \bibinfo{volume}{12}, \bibinfo{number}{7} (\bibinfo{year}{2023}), \bibinfo{pages}{01--18}.
\newblock
\showISSN{2079-9292}
\urldef\tempurl%
\url{https://doi.org/10.3390/electronics12071643}
\showDOI{\tempurl}


\bibitem[Lin et~al\mbox{.}(2024)]%
        {lin2023awq}
\bibfield{author}{\bibinfo{person}{Ji Lin}, \bibinfo{person}{Jiaming Tang}, \bibinfo{person}{Haotian Tang}, \bibinfo{person}{Shang Yang}, \bibinfo{person}{Wei-Ming Chen}, \bibinfo{person}{Wei-Chen Wang}, \bibinfo{person}{Guangxuan Xiao}, \bibinfo{person}{Xingyu Dang}, \bibinfo{person}{Chuang Gan}, {and} \bibinfo{person}{Song Han}.} \bibinfo{year}{2024}\natexlab{}.
\newblock \showarticletitle{AWQ: Activation-aware Weight Quantization for On-Device LLM Compression and Acceleration}. In \bibinfo{booktitle}{\emph{Proceedings of Machine Learning and Systems}}, \bibfield{editor}{\bibinfo{person}{P.~Gibbons}, \bibinfo{person}{G.~Pekhimenko}, {and} \bibinfo{person}{C.~De Sa}} (Eds.), Vol.~\bibinfo{volume}{6}. \bibinfo{publisher}{proceedings.mlsys.org}, \bibinfo{address}{Convention Center Drive, Miami Beach}, \bibinfo{pages}{87--100}.
\newblock
\urldef\tempurl%
\url{https://proceedings.mlsys.org/paper_files/paper/2024/file/42a452cbafa9dd64e9ba4aa95cc1ef21-Paper-Conference.pdf}
\showURL{%
\tempurl}


\bibitem[Liu et~al\mbox{.}(2020)]%
        {liu2020multi}
\bibfield{author}{\bibinfo{person}{Wenbin Liu}, \bibinfo{person}{Bojian Wen}, \bibinfo{person}{Shang Gao}, \bibinfo{person}{Jiesheng Zheng}, {and} \bibinfo{person}{Yinlong Zheng}.} \bibinfo{year}{2020}\natexlab{}.
\newblock \showarticletitle{A multi-label text classification model based on ELMo and attention}. In \bibinfo{booktitle}{\emph{MATEC Web of Conferences}}, Vol.~\bibinfo{volume}{309}. \bibinfo{publisher}{EDP Sciences}, \bibinfo{address}{Online}, \bibinfo{pages}{03015}.
\newblock


\bibitem[Liu et~al\mbox{.}(2019)]%
        {yinhan19}
\bibfield{author}{\bibinfo{person}{Y. Liu}, \bibinfo{person}{M. Ott}, \bibinfo{person}{N. Goyal}, \bibinfo{person}{J. Du}, \bibinfo{person}{M. Joshi}, \bibinfo{person}{D. Chen}, \bibinfo{person}{O. Levy}, \bibinfo{person}{M. Lewis}, \bibinfo{person}{L. Zettlemoyer}, {and} \bibinfo{person}{V. Stoyanov}.} \bibinfo{year}{2019}\natexlab{}.
\newblock \bibinfo{title}{{RoBERTa}: A Robustly Optimized {BERT} Pretraining Approach}.
\newblock
\newblock
\urldef\tempurl%
\url{https://arxiv.org/abs/1907.11692}
\showURL{%
\tempurl}


\bibitem[Lundberg et~al\mbox{.}(2018)]%
        {lundberg18}
\bibfield{author}{\bibinfo{person}{S.~M. Lundberg}, \bibinfo{person}{G.~G. Erion}, {and} \bibinfo{person}{S.-I. Lee}.} \bibinfo{year}{2018}\natexlab{}.
\newblock \bibinfo{title}{Consistent Individualized Feature Attribution for Tree Ensembles}.
\newblock
\newblock
\urldef\tempurl%
\url{https://arxiv.org/abs/1802.03888}
\showURL{%
\tempurl}


\bibitem[McInnes et~al\mbox{.}(2018)]%
        {mcinnes18}
\bibfield{author}{\bibinfo{person}{L. McInnes}, \bibinfo{person}{J. Healy}, {and} \bibinfo{person}{J. Melville}.} \bibinfo{year}{2018}\natexlab{}.
\newblock \bibinfo{title}{{UMAP}: Uniform Manifold Approximation and Projection for Dimension Reduction}.
\newblock
\newblock
\urldef\tempurl%
\url{https://arxiv.org/abs/1802.03426}
\showURL{%
\tempurl}


\bibitem[Minakawa et~al\mbox{.}(2022)]%
        {10.1145/3487553.3524643}
\bibfield{author}{\bibinfo{person}{Naoto Minakawa}, \bibinfo{person}{Kiyoshi Izumi}, \bibinfo{person}{Hiroki Sakaji}, {and} \bibinfo{person}{Hitomi Sano}.} \bibinfo{year}{2022}\natexlab{}.
\newblock \showarticletitle{Graph Representation Learning of Banking Transaction Network with Edge Weight-Enhanced Attention and Textual Information}. In \bibinfo{booktitle}{\emph{Companion Proceedings of the Web Conference 2022}} (Virtual Event, Lyon, France) \emph{(\bibinfo{series}{WWW '22})}. \bibinfo{publisher}{Association for Computing Machinery}, \bibinfo{address}{New York, NY, USA}, \bibinfo{pages}{630–637}.
\newblock
\showISBNx{9781450391306}
\urldef\tempurl%
\url{https://doi.org/10.1145/3487553.3524643}
\showDOI{\tempurl}


\bibitem[Paszke et~al\mbox{.}(2017)]%
        {pytorch}
\bibfield{author}{\bibinfo{person}{Adam Paszke}, \bibinfo{person}{Sam Gross}, \bibinfo{person}{Soumith Chintala}, \bibinfo{person}{Gregory Chanan}, \bibinfo{person}{Edward Yang}, \bibinfo{person}{Zach DeVito}, \bibinfo{person}{Zeming Lin}, \bibinfo{person}{Alban Desmaison}, \bibinfo{person}{Luca Antiga}, {and} \bibinfo{person}{Adam Lerer}.} \bibinfo{year}{2017}\natexlab{}.
\newblock \showarticletitle{Automatic differentiation in PyTorch}. In \bibinfo{booktitle}{\emph{31st Conference on Neural Information Processing Systems}}. \bibinfo{publisher}{ACM Digital Library}, \bibinfo{address}{Long Beach, CA, USA}, \bibinfo{pages}{01--04}.
\newblock
\urldef\tempurl%
\url{https://api.semanticscholar.org/CorpusID:40027675}
\showURL{%
\tempurl}


\bibitem[Pedregosa et~al\mbox{.}(2011)]%
        {scikit-learn}
\bibfield{author}{\bibinfo{person}{F. Pedregosa}, \bibinfo{person}{G. Varoquaux}, \bibinfo{person}{A. Gramfort}, \bibinfo{person}{V. Michel}, \bibinfo{person}{B. Thirion}, \bibinfo{person}{O. Grisel}, \bibinfo{person}{M. Blondel}, \bibinfo{person}{P. Prettenhofer}, \bibinfo{person}{R. Weiss}, \bibinfo{person}{V. Dubourg}, \bibinfo{person}{J. Vanderplas}, \bibinfo{person}{A. Passos}, \bibinfo{person}{D. Cournapeau}, \bibinfo{person}{M. Brucher}, \bibinfo{person}{M. Perrot}, {and} \bibinfo{person}{E. Duchesnay}.} \bibinfo{year}{2011}\natexlab{}.
\newblock \showarticletitle{Scikit-learn: Machine Learning in {P}ython}.
\newblock \bibinfo{journal}{\emph{Journal of Machine Learning Research}}  \bibinfo{volume}{12} (\bibinfo{year}{2011}), \bibinfo{pages}{2825--2830}.
\newblock


\bibitem[Peters et~al\mbox{.}(2018)]%
        {peters18}
\bibfield{author}{\bibinfo{person}{Matthew~E. Peters}, \bibinfo{person}{Mark Neumann}, \bibinfo{person}{Mohit Iyyer}, \bibinfo{person}{Matt Gardner}, \bibinfo{person}{Christopher Clark}, \bibinfo{person}{Kenton Lee}, {and} \bibinfo{person}{Luke Zettlemoyer}.} \bibinfo{year}{2018}\natexlab{}.
\newblock \showarticletitle{Deep Contextualized Word Representations}.
\newblock \bibinfo{journal}{\emph{ArXiv}}  \bibinfo{volume}{abs/1802.05365} (\bibinfo{year}{2018}), \bibinfo{pages}{01--15}.
\newblock
\urldef\tempurl%
\url{https://api.semanticscholar.org/CorpusID:3626819}
\showURL{%
\tempurl}


\bibitem[Pun and Lawryshyn(2012)]%
        {pun12}
\bibfield{author}{\bibinfo{person}{J. Pun} {and} \bibinfo{person}{Y. Lawryshyn}.} \bibinfo{year}{2012}\natexlab{}.
\newblock \showarticletitle{Improving Credit Card Fraud Detection Using a Meta-Classification Strategy}.
\newblock \bibinfo{journal}{\emph{International Journal of Computer Applications}} \bibinfo{volume}{56}, \bibinfo{number}{10} (\bibinfo{year}{2012}), \bibinfo{pages}{41--46}.
\newblock


\bibitem[Sak et~al\mbox{.}(2014)]%
        {sak2014}
\bibfield{author}{\bibinfo{person}{H. Sak}, \bibinfo{person}{A. Senior}, {and} \bibinfo{person}{F. Beaufays}.} \bibinfo{year}{2014}\natexlab{}.
\newblock \bibinfo{title}{Long Short-term Memory Based Recurrent Neural Network Architectures for Large Vocabulary Speech Recognition}.
\newblock
\newblock
\urldef\tempurl%
\url{https://arxiv.org/abs/1402.1128}
\showURL{%
\tempurl}


\bibitem[Sanh et~al\mbox{.}(2019)]%
        {sanh18}
\bibfield{author}{\bibinfo{person}{V. Sanh}, \bibinfo{person}{L. Debut}, \bibinfo{person}{J. Chaumond}, {and} \bibinfo{person}{T. Wolf}.} \bibinfo{year}{2019}\natexlab{}.
\newblock \bibinfo{title}{{DistilBERT}, a Distilled Version of {BERT}: Smaller, Faster, Cheaper and Lighter}.
\newblock
\newblock
\urldef\tempurl%
\url{https://arxiv.org/abs/1910.01108}
\showURL{%
\tempurl}


\bibitem[Schuster and Paliwal(1997)]%
        {schuster97}
\bibfield{author}{\bibinfo{person}{M. Schuster} {and} \bibinfo{person}{K.~K. Paliwal}.} \bibinfo{year}{1997}\natexlab{}.
\newblock \showarticletitle{Bidirectional Recurrent Neural Networks}.
\newblock \bibinfo{journal}{\emph{IEEE Transactions on Signal Processing}} \bibinfo{volume}{45}, \bibinfo{number}{11} (\bibinfo{year}{1997}), \bibinfo{pages}{2673--2681}.
\newblock


\bibitem[Sennrich et~al\mbox{.}(2016)]%
        {sennrich16}
\bibfield{author}{\bibinfo{person}{Rico Sennrich}, \bibinfo{person}{Barry Haddow}, {and} \bibinfo{person}{Alexandra Birch}.} \bibinfo{year}{2016}\natexlab{}.
\newblock \showarticletitle{Neural Machine Translation of Rare Words with Subword Units}. In \bibinfo{booktitle}{\emph{Proceedings of the 54th Annual Meeting of the Association for Computational Linguistics (Volume 1: Long Papers)}}, \bibfield{editor}{\bibinfo{person}{Katrin Erk} {and} \bibinfo{person}{Noah~A. Smith}} (Eds.). \bibinfo{publisher}{Association for Computational Linguistics}, \bibinfo{address}{Berlin, Germany}, \bibinfo{pages}{1715--1725}.
\newblock
\urldef\tempurl%
\url{https://doi.org/10.18653/v1/P16-1162}
\showDOI{\tempurl}


\bibitem[Shaw et~al\mbox{.}(2001)]%
        {shaw01}
\bibfield{author}{\bibinfo{person}{M.~J. Shaw}, \bibinfo{person}{C. Subramaniam}, \bibinfo{person}{G.~W. Tan}, {and} \bibinfo{person}{M.~E. Welge}.} \bibinfo{year}{2001}\natexlab{}.
\newblock \showarticletitle{Knowledge Management and Data Mining for Marketing}.
\newblock \bibinfo{journal}{\emph{Decision Support Systems}} \bibinfo{volume}{31}, \bibinfo{number}{1} (\bibinfo{year}{2001}), \bibinfo{pages}{127--137}.
\newblock


\bibitem[Shchur et~al\mbox{.}(2021)]%
        {shchur2021neural}
\bibfield{author}{\bibinfo{person}{Oleksandr Shchur}, \bibinfo{person}{Ali~Caner T{\"u}rkmen}, \bibinfo{person}{Tim Januschowski}, {and} \bibinfo{person}{Stephan G{\"u}nnemann}.} \bibinfo{year}{2021}\natexlab{}.
\newblock \showarticletitle{Neural Temporal Point Processes: A Review}.
\newblock \bibinfo{journal}{\emph{ArXiv}}  \bibinfo{volume}{abs/2104.03528} (\bibinfo{year}{2021}), \bibinfo{pages}{01--09}.
\newblock
\urldef\tempurl%
\url{https://api.semanticscholar.org/CorpusID:233181707}
\showURL{%
\tempurl}


\bibitem[Smeureanu et~al\mbox{.}(2013)]%
        {smeureanu13}
\bibfield{author}{\bibinfo{person}{I. Smeureanu}, \bibinfo{person}{G. Ruxanda}, {and} \bibinfo{person}{L.~M. Badea}.} \bibinfo{year}{2013}\natexlab{}.
\newblock \showarticletitle{Customer Segmentation in Private Banking Sector Using Machine Learning Techniques}.
\newblock \bibinfo{journal}{\emph{Journal of Business Economics and Management}} \bibinfo{volume}{14}, \bibinfo{number}{5} (\bibinfo{year}{2013}), \bibinfo{pages}{923--939}.
\newblock


\bibitem[Smith and Topin(2019)]%
        {smith19}
\bibfield{author}{\bibinfo{person}{Leslie~N. Smith} {and} \bibinfo{person}{Nicholay Topin}.} \bibinfo{year}{2019}\natexlab{}.
\newblock \showarticletitle{{Super-convergence: very fast training of neural networks using large learning rates}}. In \bibinfo{booktitle}{\emph{Artificial Intelligence and Machine Learning for Multi-Domain Operations Applications}}, \bibfield{editor}{\bibinfo{person}{Tien Pham}} (Ed.), Vol.~\bibinfo{volume}{11006}. International Society for Optics and Photonics, \bibinfo{publisher}{SPIE}, \bibinfo{address}{Baltimore, MD, United States}, \bibinfo{pages}{369--386}.
\newblock
\urldef\tempurl%
\url{https://doi.org/10.1117/12.2520589}
\showDOI{\tempurl}


\bibitem[Smith(1956)]%
        {smith56}
\bibfield{author}{\bibinfo{person}{W.~R. Smith}.} \bibinfo{year}{1956}\natexlab{}.
\newblock \showarticletitle{Product Differentiation and Market Segmentation as Alternative Marketing Strategies}.
\newblock \bibinfo{journal}{\emph{Journal of Marketing}} \bibinfo{volume}{21}, \bibinfo{number}{1} (\bibinfo{year}{1956}), \bibinfo{pages}{3--8}.
\newblock


\bibitem[Vaswani et~al\mbox{.}(2017)]%
        {vaswani17}
\bibfield{author}{\bibinfo{person}{Ashish Vaswani}, \bibinfo{person}{Noam Shazeer}, \bibinfo{person}{Niki Parmar}, \bibinfo{person}{Jakob Uszkoreit}, \bibinfo{person}{Llion Jones}, \bibinfo{person}{Aidan~N Gomez}, \bibinfo{person}{\L~ukasz Kaiser}, {and} \bibinfo{person}{Illia Polosukhin}.} \bibinfo{year}{2017}\natexlab{}.
\newblock \showarticletitle{Attention is All you Need}. In \bibinfo{booktitle}{\emph{Advances in Neural Information Processing Systems}}, \bibfield{editor}{\bibinfo{person}{I.~Guyon}, \bibinfo{person}{U.~Von Luxburg}, \bibinfo{person}{S.~Bengio}, \bibinfo{person}{H.~Wallach}, \bibinfo{person}{R.~Fergus}, \bibinfo{person}{S.~Vishwanathan}, {and} \bibinfo{person}{R.~Garnett}} (Eds.), Vol.~\bibinfo{volume}{30}. \bibinfo{publisher}{Curran Associates, Inc.}, \bibinfo{address}{Long Beach, CA, USA}, \bibinfo{pages}{01--11}.
\newblock
\urldef\tempurl%
\url{https://proceedings.neurips.cc/paper_files/paper/2017/file/3f5ee243547dee91fbd053c1c4a845aa-Paper.pdf}
\showURL{%
\tempurl}


\bibitem[\v{S}trumbelj and Kononenko(2010)]%
        {strumbelj10}
\bibfield{author}{\bibinfo{person}{E. \v{S}trumbelj} {and} \bibinfo{person}{I. Kononenko}.} \bibinfo{year}{2010}\natexlab{}.
\newblock \showarticletitle{An efficient explanation of individual classifications using game theory}.
\newblock \bibinfo{journal}{\emph{The Journal of Machine Learning Research}}  \bibinfo{volume}{11} (\bibinfo{year}{2010}), \bibinfo{pages}{1--18}.
\newblock


\bibitem[Wiese and Omlin(2009)]%
        {wiese09}
\bibfield{author}{\bibinfo{person}{B. Wiese} {and} \bibinfo{person}{C. Omlin}.} \bibinfo{year}{2009}\natexlab{}.
\newblock \showarticletitle{Credit Card Transactions, Fraud Detection, and Machine Learning: Modelling Time with LSTM Recurrent Neural Networks}.
\newblock In \bibinfo{booktitle}{\emph{Innovations in neural information paradigms and applications}}. \bibinfo{publisher}{Springer}, \bibinfo{address}{Berlin, Germany}, \bibinfo{pages}{231--268}.
\newblock


\bibitem[Yan(2019)]%
        {yan2019recent}
\bibfield{author}{\bibinfo{person}{Junchi Yan}.} \bibinfo{year}{2019}\natexlab{}.
\newblock \showarticletitle{Recent advance in temporal point process: from machine learning perspective}.
\newblock \bibinfo{journal}{\emph{SJTU Technical Report}}  \bibinfo{volume}{Thinklab} (\bibinfo{year}{2019}), \bibinfo{pages}{01--07}.
\newblock


\bibitem[Zhang et~al\mbox{.}(2020)]%
        {zhang20}
\bibfield{author}{\bibinfo{person}{Jingqing Zhang}, \bibinfo{person}{Yao Zhao}, \bibinfo{person}{Mohammad Saleh}, {and} \bibinfo{person}{Peter~J. Liu}.} \bibinfo{year}{2020}\natexlab{}.
\newblock \showarticletitle{PEGASUS: pre-training with extracted gap-sentences for abstractive summarization}. In \bibinfo{booktitle}{\emph{Proceedings of the 37th International Conference on Machine Learning}} \emph{(\bibinfo{series}{ICML'20})}. \bibinfo{publisher}{JMLR.org}, \bibinfo{address}{Online}, Article \bibinfo{articleno}{1051}, \bibinfo{numpages}{12}~pages.
\newblock


\end{thebibliography}

\appendix

\counterwithin{figure}{section}
\section{Overview of the preprocessing} \label{append:prepro}

In this appendix we will illustrate all the transformations made on BTF through an example inspired by table \ref{tab:preprolib}. An example of raw data is shown on Table \ref{appendix:raw_data}. 
First, we apply normalization and ordering, as shown on Table \ref{appendix:norm_order}.\\

\begin{table}[htbp]
    \centering
    \begin{tabular}{ |c|c|c| } 
        \hline
        \textbf{date} & \textbf{amount} & \textbf{wording} \\
        \hline
        2021-09-03 & 1010 & VIR POLE EMPLOI BRETAGNE 08/21 \\
        2021-09-03 & -100 & CHQ 2141367 \\
        \hline
        2021-09-11 & -42  & CARTE 08/10 LECLERC BREST \\
        \hline
        2021-09-20 & -50 & RET DAB 351267 PLANCOET \\
        \hline
    \end{tabular}
    \caption{An example of raw data.}
    \label{appendix:raw_data}
    \vskip -0.25in
\end{table}

\begin{table}[htbp]
    \centering
    \begin{tabular}{ |c|c|c| } 
        \hline
        \textbf{date} & \textbf{amount} & \textbf{wording} \\
        \hline
        2021-09-03 & -100 & chq $<$digits$>$ \\
        2021-09-03 & 1010 & vir pole emploi bretagne $<$date$>$ \\
        \hline
        2021-09-11 & -42  & carte $<$date$>$ leclerc brest \\
        \hline
        2021-09-20 & -50 & ret dab $<$digits$>$ plancoet \\
        \hline
    \end{tabular}
    \caption{The result of normalization and ordering steps.}
    \vskip -0.25in
    \label{appendix:norm_order}
\end{table}

Finally, after adding the BOS and EOS control tokens, the sequence labels tokenization gives:\\
Wordings: [BOS] $|$ \_ $|$ chq$<$digits$>$ $|$ \_virpoleemploi $|$ bretagne $|$ $<$date$>$ $|$ \_carte$<$date$>$ $|$ leclerc $|$ brest $|$ \_retdab$<$digits$>$ $|$ plancoet $|$ [EOS]\\
We reflect the normalized day number in the month on all the wordings sub-tokens:\\
Dates: [BOS] $|$ $\frac{3}{30}$ $|$ $\frac{3}{30}$ $|$ $\frac{3}{30}$ $|$ $\frac{3}{30}$ $|$ $\frac{3}{30}$ $|$ $\frac{11}{30}$ $|$ $\frac{11}{30}$ $|$ $\frac{11}{30}$ $|$ $\frac{20}{30}$ $|$ $\frac{20}{30}$ $|$ [EOS]\\
Then to finish we do the same with the amounts:\\
Amounts: [BOS] $|$ -100$|$ -100 $|$ 1010 $|$ 1010 $|$ 1010 $|$ -42 $|$ -42 $|$ -42 $|$ -50 $|$ -50 $|$ [EOS]

\begin{figure}[ht]
    \centering
    \includegraphics[scale=0.63]{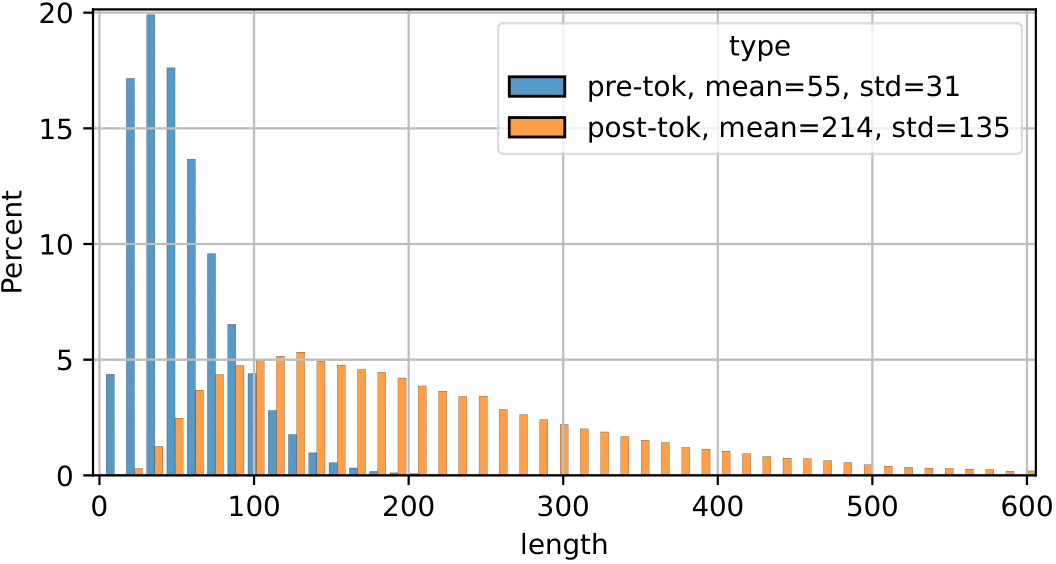}
    \vskip -0.1in
    \caption{Distribution of 10k sequences based on one month of bank transactions.}
    \label{fig:seq-length}
    \Description[<short description>]{<long description>}
    \vskip -0.1in
\end{figure}

After tokenization the length of the sequences is strongly increased. In figure \ref{fig:seq-length} we plot the distribution of sequences consisting of one month of transactions.

\section{About Pre-training} \label{append:pretrain}

\begin{figure*}[ht]
    \vskip -0.1in
    \centering
    \includegraphics[scale=0.8]{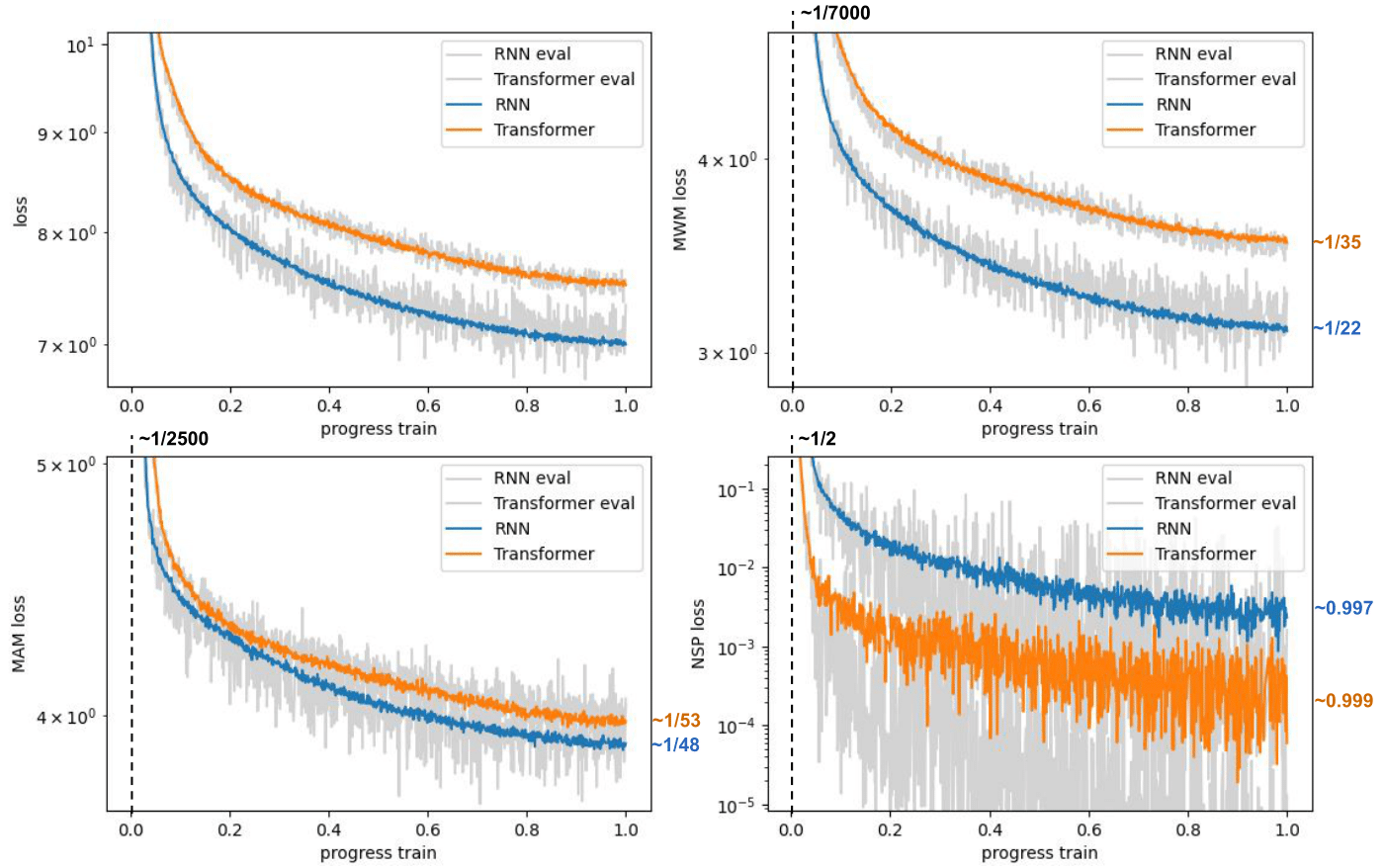}
    \vskip -0.2in
    \caption{Pre-training loss functions.}
    \label{fig:train-curve}
    \Description[<short description>]{<long description>}
    \vskip -0.15in
\end{figure*}

For the pre-training, the Pytorch \cite{pytorch} backend is used and two additional control tokens are added: the masking token [MASK] whose purpose is to mask an event in the wordings or amounts sequence and a padding token [PAD] allowing to create a mini-batch with different sequence sizes. Thanks to this token the attention mechanism will either not be taken into account for the Transformer architecture or simply not calculated for the RNN.\\
It is also important to note that the classification token of a sequence is the embeeded representation of the [BOS] token for Transformer. However, in order to take into account the specificities of the bi-directional RNN structure for RNN modeling, the embedded classification vector is the sum of the embeeding representations of the tokens [BOS] and [EOS].\\
In figure \ref{fig:train-curve} we can observe the evolution of the loss functions of each pre-training subtasks. The values indicated are the probabilities of finding the right answers for each subtasks.

\section{More Details on Downstream Tasks}

In this appendix we will discuss in more detail the tasks tested for the performance evaluation of the models.

\subsection{Transaction Classification} \label{append:transclassif}

\begin{figure}[ht]
    \centering
    \includegraphics[scale=1]{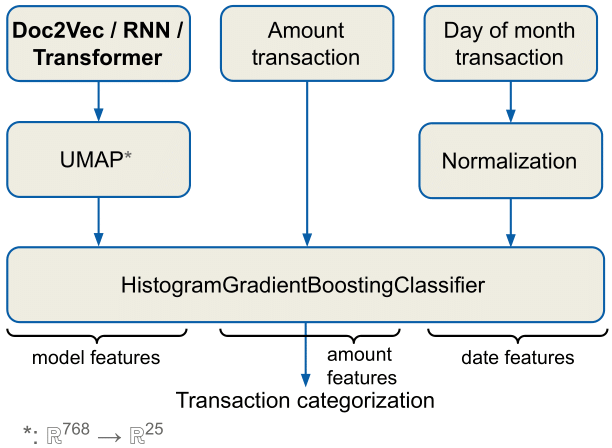}
    \vskip -0.15in
    \caption{Test structure of the different models for the transaction classification task.}
    \label{fig:transaction-pred}
    \Description[<short description>]{<long description>}
    \vskip -0.15in
\end{figure}

\begin{figure*}[p]
    \centering
    \begin{subfigure}[b]{0.38\textwidth}
        \centering
        \includegraphics[width=\textwidth]{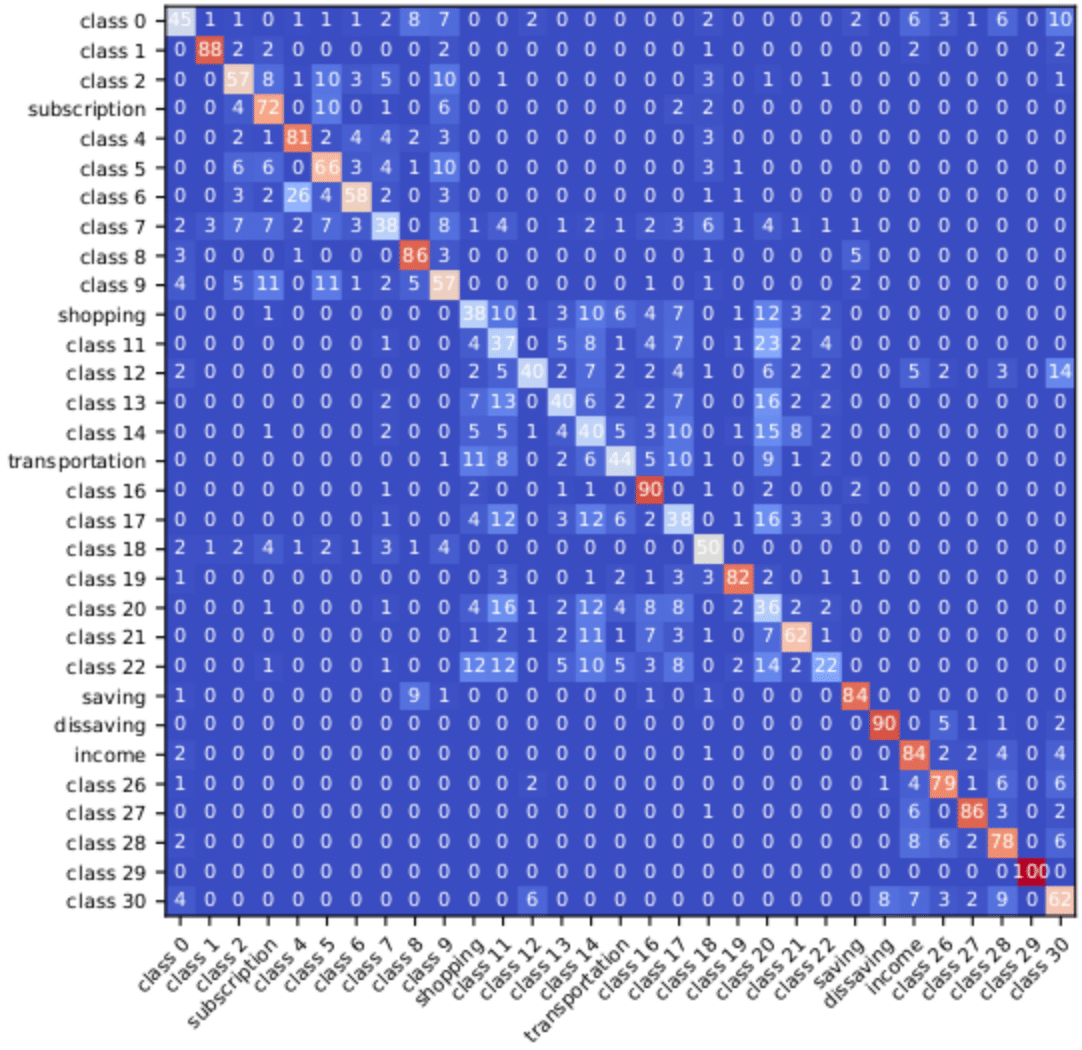}
        \vskip -0.09in
        \caption{Doc2Vec}
    \end{subfigure}
    \begin{subfigure}[b]{0.38\textwidth}
        \centering
        \includegraphics[width=\textwidth]{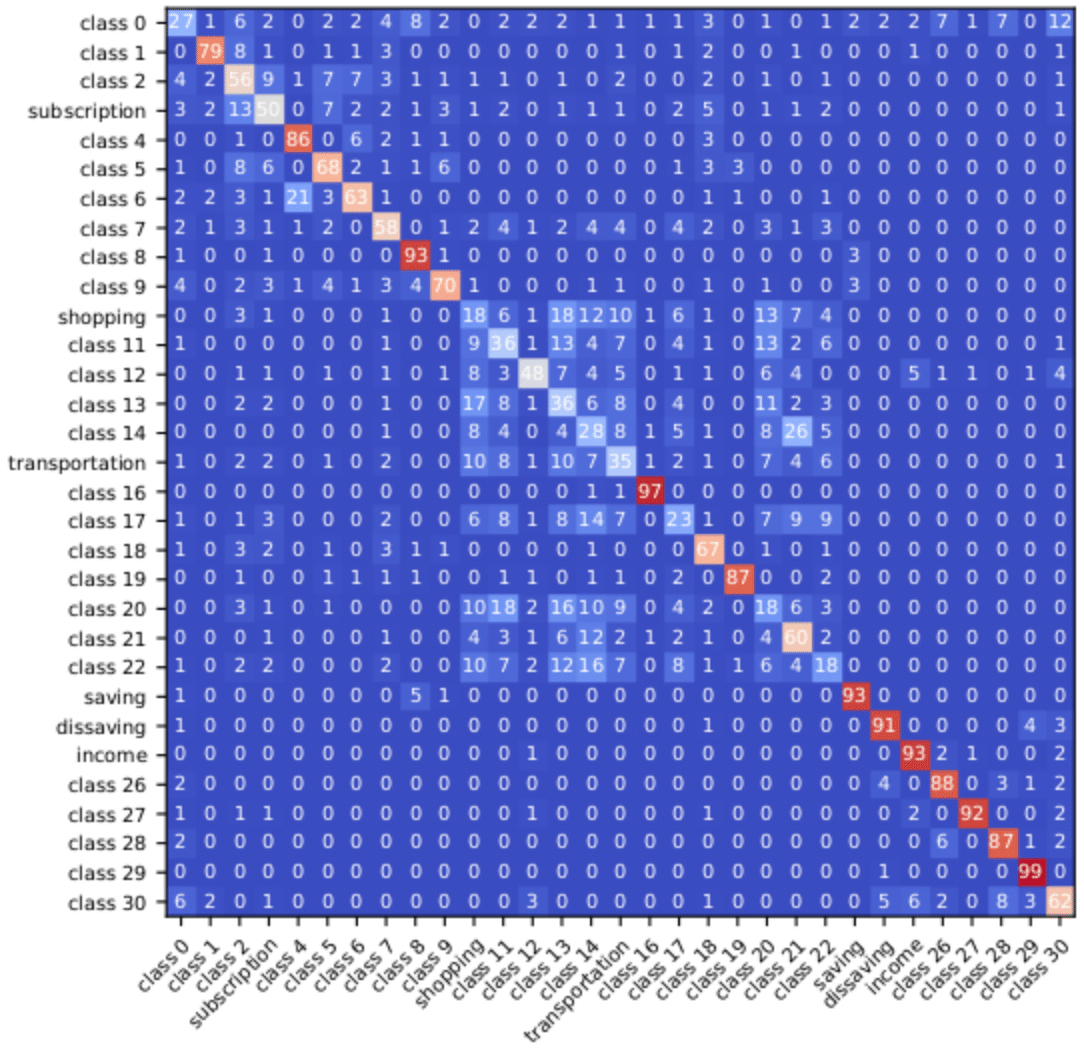}
        \vskip -0.09in
        \caption{RNN}
    \end{subfigure}
    \begin{subfigure}[b]{0.38\textwidth}
        \centering
        \includegraphics[width=\textwidth]{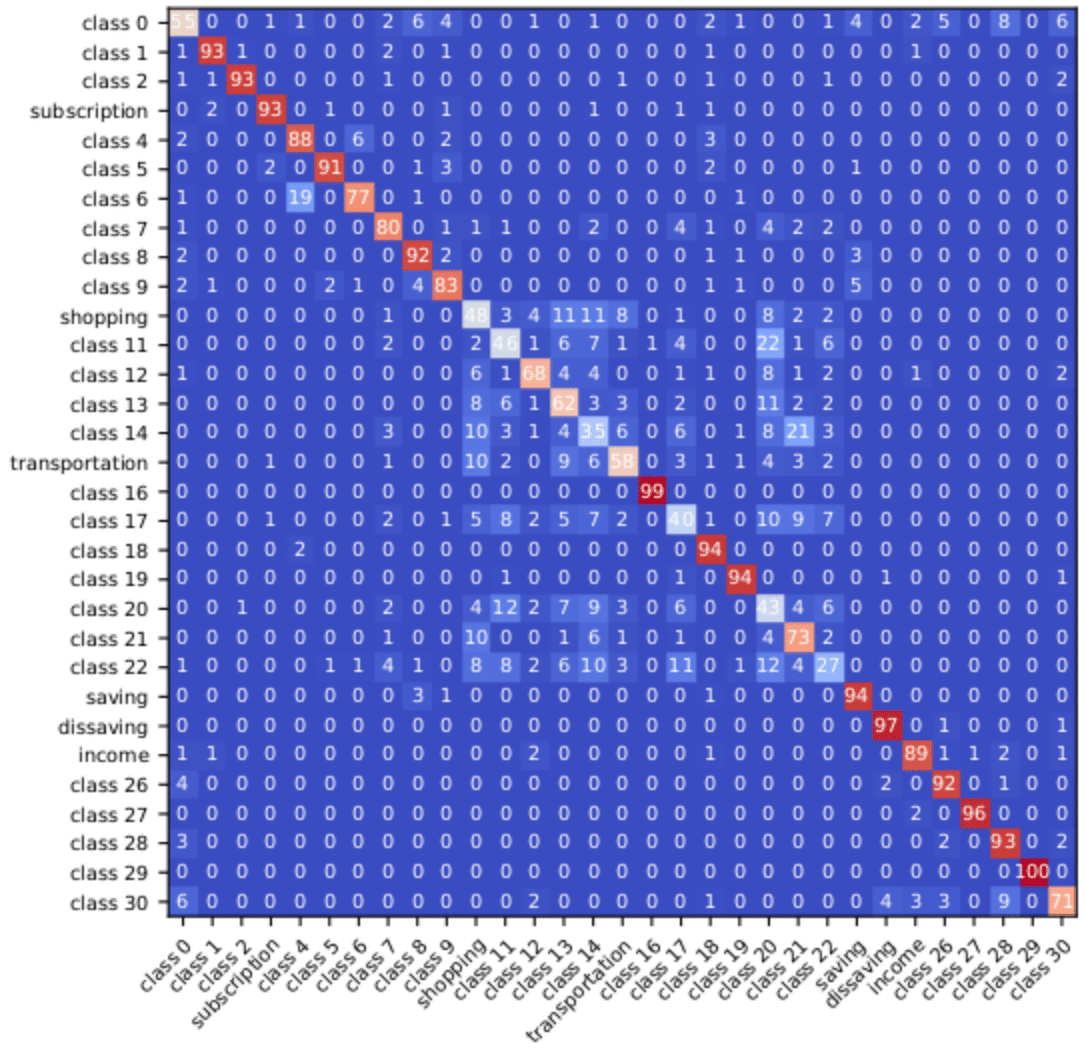}
        \vskip -0.09in
        \caption{Transformer}
    \end{subfigure}
    \begin{subfigure}[b]{0.38\textwidth}
        \centering
        \includegraphics[width=\textwidth]{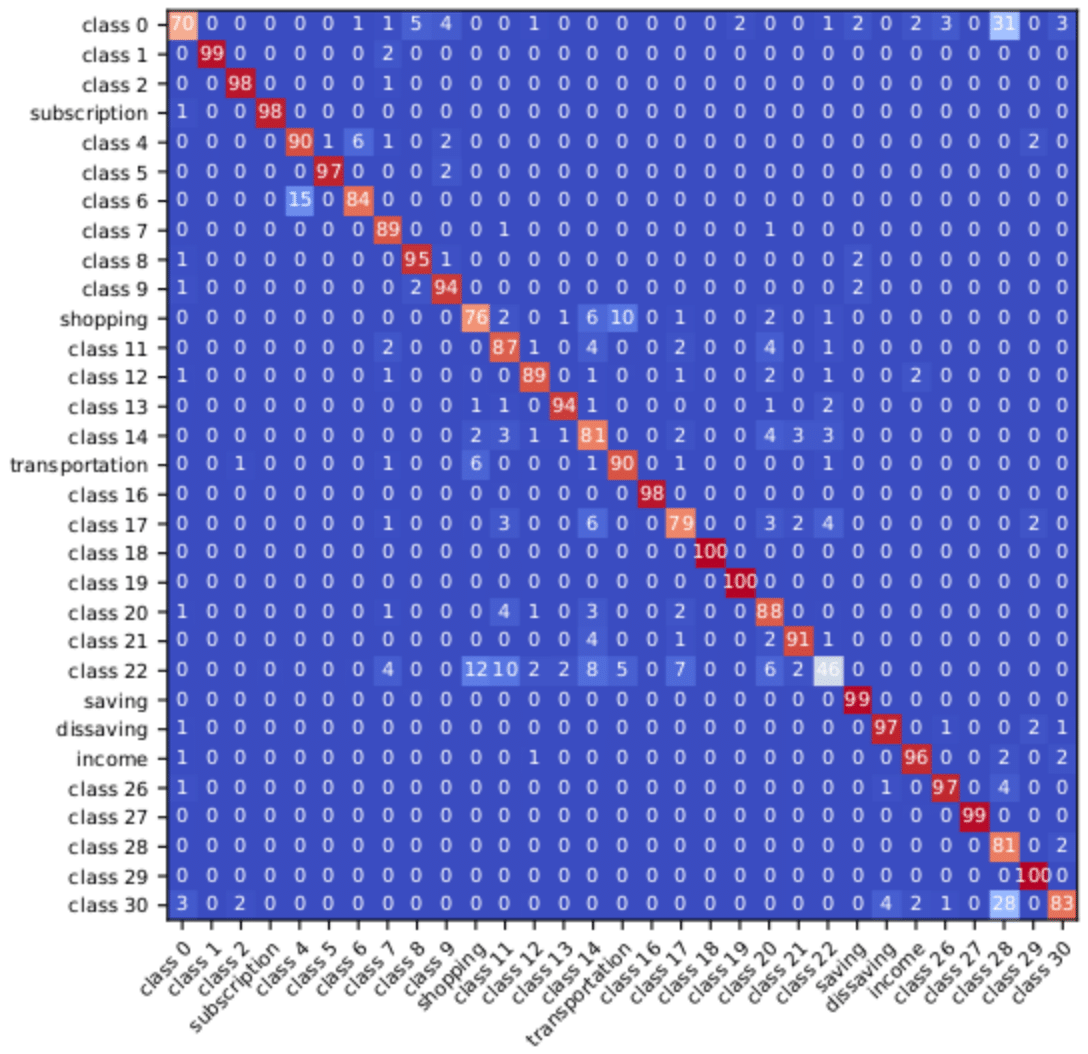}
        \vskip -0.09in
        \caption{RNN finetuned}
    \end{subfigure}
    \begin{subfigure}[b]{0.38\textwidth}
        \centering
        \includegraphics[width=\textwidth]{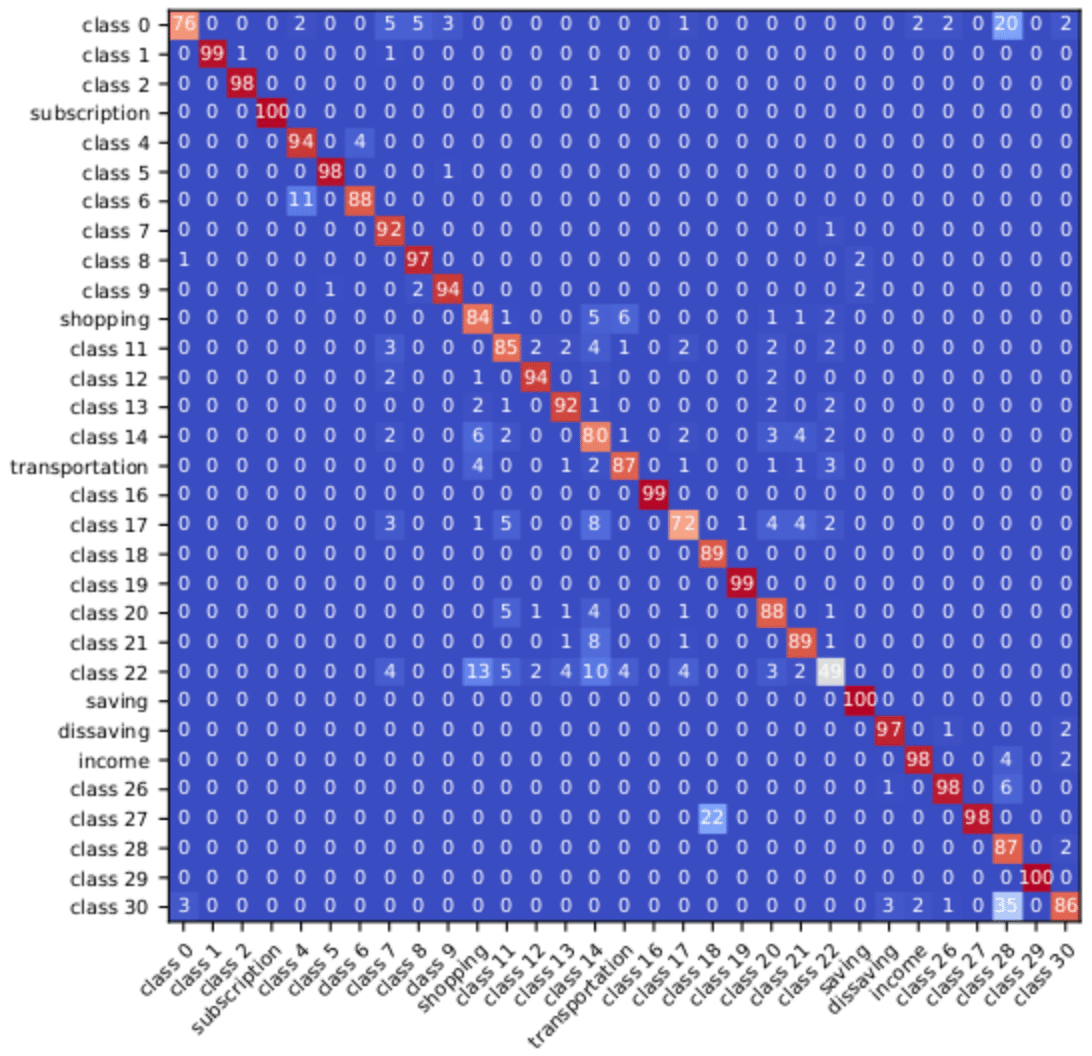}
        \vskip -0.09in
        \caption{Transformer finetuned}
    \end{subfigure}
    \vskip -0.15in
    \caption{Confusion matrix of transactions categorization task.}
    \label{fig:mc}
    \Description[<short description>]{<long description>}
\end{figure*}

We will voluntarily not enumerate the categories. This is a 31-class classification problem and the classes are for example: ``income", ``shopping", ``subscription", ``transportation" (gas, transit, repair), ``savings", ``dissaving", \textit{etc.} The training dataset is composed in a such way that each transaction category is present at least (if possible) 1.6k times in different sequences. As for the evaluation dataset, it is composed of 400 observations per category (if possible) contained in different sequences. As a reminder, the sequence is composed of 2 months of banking transactions.\\
In order to not making the features number too disproportionate between the different feature topologies, we have incorporated a non-linear feature extraction technique called Uniform Manifold Approximation and Projection (UMAP) \cite{mcinnes18} at the output of Doc2Vec, RNN and Transformer models which permits to ``reduce" the dimensions from 768 to 25.\\
For GBDT, the Scikit-Learn HistogramGradientBoostingClassifier \cite{scikit-learn} implementation was chosen and the hyperparameters set was chosen after a search for optimal hyperparameters by cross-validation.\\
Figure \ref{fig:transaction-pred} illustrates the used test structure. The amount transaction and the day of the month are both simultaneously given as an input of the RNN and Transformer models, as well as an input of the GBDT. Their are not given to the Doc2Vec. The wording of the transaction is given to the Doc2Vec, the RNN and the Transformer models, but not to the GBDT. This specific structure allows to quantify in which extend the GBDT prediction lies on the pre-trained models, by using Shapley values.\\
The confusion matrices of each model can be found on Figure \ref{fig:mc}.

\subsection{Credit Risk Scoring} \label{append:creditrisk}

This depends on the banking company nature, as some are specialized in subprime loans. But in most cases it is important to note that serious repayment defaults are very rare events. Therefore, after labeling by the definition given for an incidental contract, we apply a downsampling of negative cases in order to have a balanced dataset. Thus our training dataset is made of 6.4k observations having as many negative cases as positive cases and our evaluation dataset has 1.6k cases also balanced.

\begin{figure*}[p]
    \centering
    \begin{subfigure}[b]{0.43\textwidth}
        \centering
        \includegraphics[width=\textwidth]{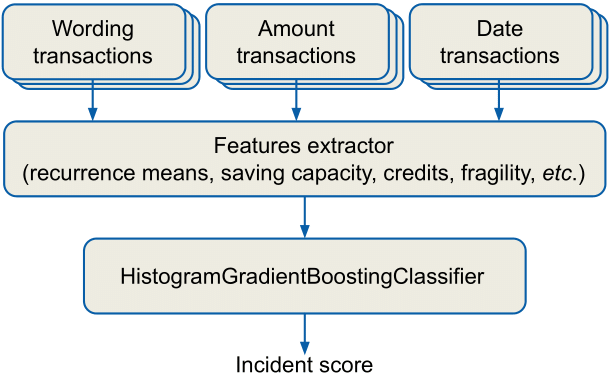}
        \caption{Reference model structure.}
        \label{fig:ref-gbdt}
    \end{subfigure}
    \begin{subfigure}[b]{0.43\textwidth}
        \centering
        \includegraphics[width=\textwidth]{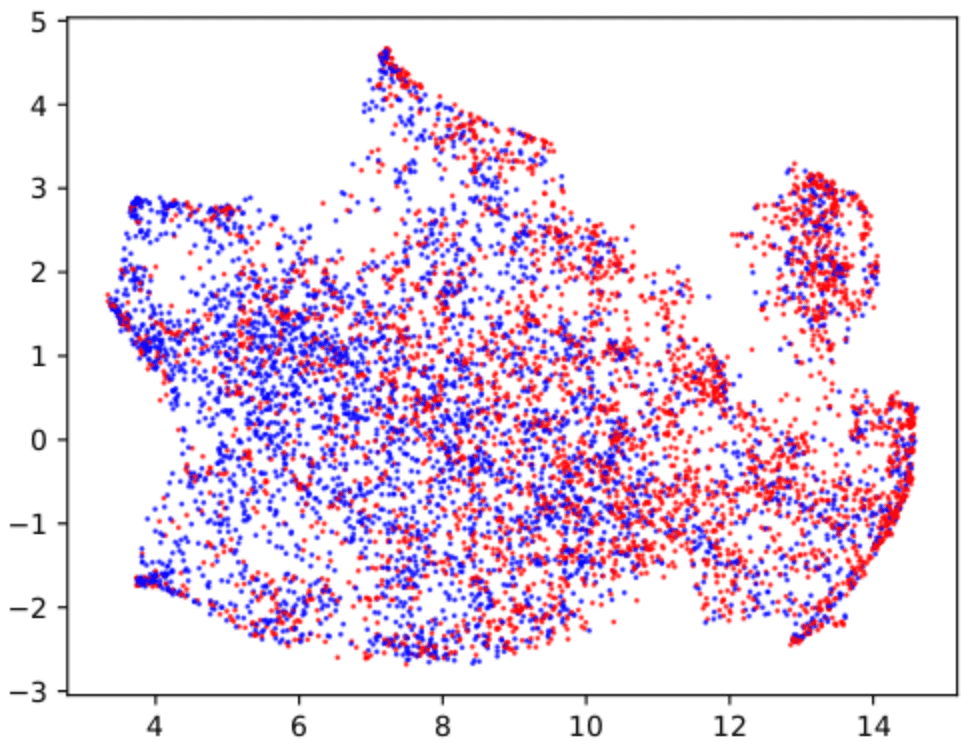}
        \caption{UMAP embedding with 18 to 2 dimensions projection, blue (resp. red) dots are negative (resp. postive) cases.}
        \label{fig:umap}
    \end{subfigure}
    \vskip -0.15in
    \caption{Information related to the reference model.}
    \label{fig:ref-risk}
    \Description[<short description>]{<long description>}
\end{figure*}

The reference modeling (fig.\ref{fig:ref-risk}\subref{fig:ref-gbdt}) consists in extracting information from the banking flow. We will not go into detail in the transformations performed during this phase, but it consists in extracting 18 features. However, we can see in figure \ref{fig:ref-risk}\subref{fig:umap} the relatively good separability of the two classes after dimensions reduction via UMAP. As shown in table \ref{tab:perf-risk} the performances of this modeling are very close to the state-of-the-art without using socio-demographic data or balances proving the transformations relevance.

\begin{figure*}[ht]
    \centering
    \includegraphics[scale=1]{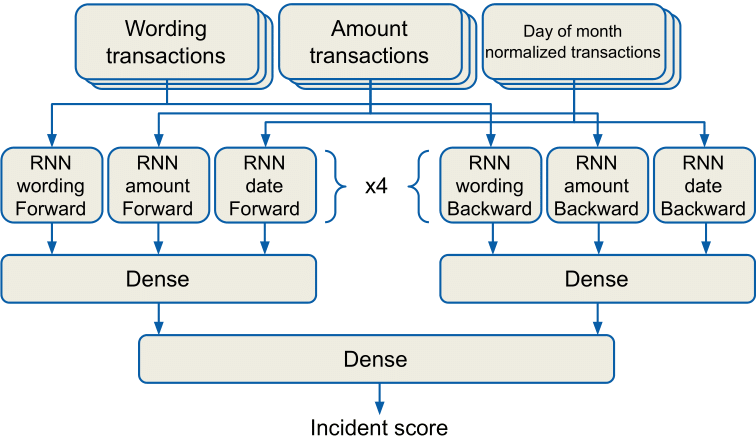}
    \vskip -0.15in
    \caption{Deep Learning reference structure.}
    \label{fig:risk-rnn}
    \Description[<short description>]{<long description>}
\end{figure*}

In order to illustrate the difficulty of directly exploiting BTF, an architecture has been developed for this sole purpose and is presented in figure \ref{fig:risk-rnn}. It consists in 4 layers of 3 bi-directional LSTMs for each modality, the embedded representations sizes of the LSTMs are 80. All these choices were made after a hyperparameter search as for the previous task. This architecture allows to jointly exploit all the modalities.

\begin{figure*}[ht]
    \centering
    \includegraphics[scale=0.7]{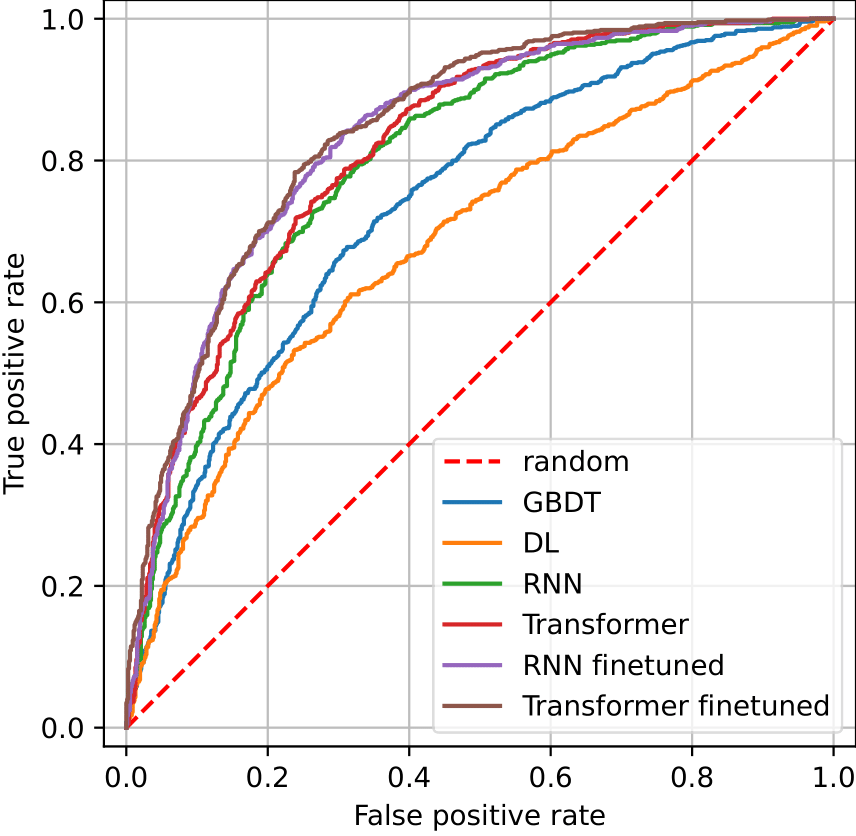}
    \vskip -0.15in
    \caption{ROC curve for credit risk task.}
    \label{fig:roc}
    \Description[<short description>]{<long description>}
    \vskip -0.15in
\end{figure*}

Finally, figure \ref{fig:roc} illustrates the ROC-AUC performances presented in table \ref{tab:perf-risk}.

\end{document}